\documentclass{article}

\usepackage{PRIMEarxiv}

\usepackage[utf8]{inputenc} 
\usepackage[T1]{fontenc}    
\usepackage{hyperref}       
\usepackage{url}            
\usepackage{booktabs}       
\usepackage{amsfonts}       
\usepackage{nicefrac}       
\usepackage{microtype}      
\usepackage{lipsum}
\usepackage{fancyhdr}       
\usepackage{graphicx}
\usepackage{amsmath}
\usepackage{amsfonts}
\usepackage{amssymb}
\usepackage{graphicx}
\usepackage{appendix}
\usepackage [english]{babel}
\usepackage [autostyle, english = american]{csquotes}
\usepackage{booktabs}
\usepackage{multirow}
\MakeOuterQuote{"}
\usepackage{setspace}
\graphicspath{{images/}}
\usepackage{amsthm}
\usepackage{array}
\usepackage{cite}

\hypersetup{hidelinks}
\usepackage{longtable}
\usepackage{fancyhdr}
\usepackage{titletoc}
\usepackage{subcaption}
\usepackage{tabularx} 
 
\usepackage{float}
\usepackage{tablefootnote}
\usepackage{caption}
\usepackage{multirow}
\usepackage{xcolor}
\usepackage{textcomp}

\graphicspath{{media/}}

\title{Augmented Regression Models using Neurochaos Learning 

}

\author{
  Akhila Henry \\
  Department of Mathematics\\
  Amrita Vishwa Vidyapeetham\\
  Amritapuri Campus \\
  Kollam, 690525, Kerala, India.\\
  \texttt{akhilahenryu@am.amrita.edu} \\
   \And
  Nithin Nagaraj\\
  Complex Systems Programme\\ National Institute of Advanced Studies\\ Indian Institute of Science Campus\\ Bengaluru, 560012,
Karnataka, India.\\
  \texttt{nithin@nias.res.in} \\
}

\begin{document}
\maketitle

\begin{abstract}
This study presents novel Augmented Regression Models using Neurochaos Learning (NL), where {\it Tracemean} features derived from the Neurochaos Learning framework are integrated with traditional regression algorithms —- {\it Linear Regression, Ridge Regression, Lasso Regression}, and {\it Support Vector Regression (SVR)}. Our approach was evaluated using ten diverse real-life datasets and a synthetically generated dataset of the form $y=mx+c+\epsilon$. Results show that incorporating the {\it Tracemean} feature (mean of the chaotic neural traces of the neurons in the NL architecture) significantly enhances regression performance, particularly in Augmented Lasso Regression and Augmented SVR, where six out of ten real-life datasets exhibited improved predictive accuracy. Among the models, Augmented Chaotic Ridge Regression achieved the highest average performance boost (11.35\%). Additionally, experiments on the simulated dataset demonstrated that the Mean Squared Error (MSE) of the augmented models consistently decreased and converged towards the Minimum Mean Squared Error (MMSE) as the sample size increased. This work demonstrates the potential of chaos-inspired features in regression tasks, offering a pathway to more accurate and computationally efficient prediction models.
\end{abstract}


\section{Introduction}

In today’s world, we are surrounded by an abundance of data generated from various domains, including healthcare, finance, agriculture, and more. Extracting meaningful insights from this data has the potential to bring transformative changes. Among the most critical tasks in data science is prediction—identifying the underlying function or model that governs data generation. Accurate prediction has far-reaching applications, from weather forecasting to sales prediction and marketing trends, directly impacting human life. For instance, precise weather forecasts can guide farmers in making informed agricultural decisions, enhancing crop yield and sustainability.

Regression, a fundamental task in machine learning, is dedicated to predicting continuous values based on input features. Numerous regression algorithms have been developed, including Linear Regression, Ridge Regression, Lasso Regression, and Support Vector Regression. These traditional methods are well-established but are often limited by their reliance on strict mathematical assumptions or their sensitivity to data distribution. To explore beyond these limitations, researchers have continuously sought novel approaches to improve prediction performance.

One such innovative approach is Neurochaos Learning (NL)~\cite{Deeksha,harikrishnan2019novel,harikrishnan2022classification}, a brain-inspired algorithm initially proposed for classification tasks. NL is characterized by transforming the original features of a dataset into chaotic features using chaotic maps, specifically the Skew Tent Map. These chaotic features—firing time, firing rate, energy, and entropy—are then used for classification by employing either cosine similarity or traditional machine learning algorithms. The success of NL in classification has inspired the exploration of its potential in regression tasks, an extension given that classification can be viewed as a specialized form of regression that predicts discrete labels instead of continuous values.

This study aims to extend the Neurochaos Learning framework to regression problems. The approach utilises a single chaotic feaure - Tracemean~\cite{henry2025neurochaos} and the original features of dataset which is then passed on to traditional regression algorithms including Linear Regression, Ridge Regression, Lasso Regression, and Support Vector Regression~\cite{fahrmeir2022regression, ma2003accurate}. By augmenting the original features with this single chaotic feature, we maintain the advantages of chaos-inspired transformations while significantly reducing computation time.

The proposed Neurochaos Learning-based regression algorithm is then evaluated using ten real life datasets : Diabetes~\cite{lichman2013uci}, Bike Sharing~\cite{bike_sharing_275}, Fish~\cite{mark_daniel_lampa_rose_claire_librojo_mary_mae_calamba_2022}, Life Expectancy~\cite{lasha_gochiashvili_2023}, Concrete Strength~\cite{concrete_compressive_strength_165}, World Happiness Report~\cite{world_happiness_kaggle}, Body Fat~\cite{body_fat_kaggle}, Auto MPG~\cite{auto_mpg_9}, Red Wine Quality~\cite{wine_quality_186}, NASA Airfoil Self-Noise~\cite{airfoil_self-noise_291}  and generated dataset of the form $\{(x_i,y_i):y_i=mx_i+c+\epsilon,i=1,2,\ldots,n\}$.  The performance is compared against existing classical methods using standard performance metrics such as R² score, Mean Squared Error (MSE), and Mean Absolute Error (MAE). This exploration not only demonstrates the versatility of NL but also highlights the potential of chaotic transformations in enhancing regression models, paving the way for more accurate and computationally efficient prediction frameworks.

\section{Traditional Regression Algorithms}
Traditional regression algorithms form the foundational methods in supervised learning used to model the relationship between input variables and a continuous output. These techniques are often effective and serve as a benchmark for evaluating newer models.
\subsection{Linear Regression}
Linear Regression ~\cite{fahrmeir2022regression} is one of the simplest and most widely used regression techniques. It models the relationship between a dependent variable and one or more independent variables by fitting a linear equation to the observed data.

Let $\{(y_i,x_{i1},\ldots,x_{ik}), i=1,2,\ldots n\}$ be the dataset , where $y$ is a continuous variable and $x_{i1},\ldots,x_{ik}$ are continuous or appropriately coded categorical regressors.
The regression model is given by 
\[
y_i = \beta_0 + \beta_1 x_{i1} + \cdots + \beta_k x_{ik} + \varepsilon_i, \quad i = 1, \ldots, n,
\]
where the error terms $\varepsilon_1, \ldots, \varepsilon_n$ are assumed to be independent and identically distributed (i.i.d.) with mean zero and constant variance:
\[
E(\varepsilon_i) = 0, \quad \text{Var}(\varepsilon_i) = \sigma^2.
\]
The coefficients $\beta_0,\beta_1,\ldots,\beta_k$ are called as regression coefficients.
\noindent
The estimated regression function is given by:
\[
\hat{f}(x_1, \ldots, x_k) = \hat{\beta}_0 + \hat{\beta}_1 x_1 + \cdots + \hat{\beta}_k x_k,
\]
which serves as an estimator for the conditional expectation $E(y \mid x_1, \ldots, x_k)$. Thus, this function can be used for predicting the response variable $y$, commonly denoted by $\hat{y}$.

\noindent
For the purposes of statistical inference, it is often further assumed that the error terms follow a normal distribution:
\[
\varepsilon_i \sim \mathcal{N}(0, \sigma^2).\]

\noindent
The estimation theory for regression coefficients in linear models is fundamentally based on the method of least squares(LS). While least squares is a widely used technique, other approaches such as maximum likelihood estimation can also be employed for parameter estimation.

\noindent
According to the least squares principle, the regression coefficients $\boldsymbol{\beta}= (\beta_0,\beta_1,\ldots,\beta_k)$ are estimated by minimizing the sum of squared residuals:
\[
LS(\boldsymbol{\beta}) = \sum_{i=1}^{n} (y_i - \mathbf{x}_i^\top \boldsymbol{\beta})^2 = \sum_{i=1}^{n} \varepsilon_i^2 = \boldsymbol{\varepsilon}^\top \boldsymbol{\varepsilon},
\]
where $\mathbf{x}_i$ is the vector of predictors for the $i^{th}$ observation and $\varepsilon_i$ is the residual error.

\subsection{Ridge Regression}
Ridge regression~\cite{fahrmeir2022regression} addresses overfitting in linear regression by introducing a regularization technique that modifies the cost function with a penalty term. This penalty, proportional to the sum of the squared coefficients, constrains their magnitude. As the penalty strength increases, the coefficients are shrunk closer to zero, thereby reducing the model's variance and helping to generalize better on unseen data.
\noindent
The penalty term, calculated as the sum of squared coefficients, discourages large parameter values and leads to a modified objective function of the form:
$$\text{LS}(\boldsymbol{\beta}) = (\mathbf{y} - \mathbf{X}\boldsymbol{\beta})^T(\mathbf{y} - \mathbf{X}\boldsymbol{\beta}) + \lambda \boldsymbol{\beta}^T \boldsymbol{\beta},$$
\noindent
where $\lambda$ controls the strength of regularization. By minimizing this penalized loss, ridge regression ensures a balance between fitting the data and maintaining model simplicity.

\subsection{Lasso Regression}
While ridge regression is effective for estimating regression coefficients in high-dimensional settings or when the features exhibits multicollinearity, it does not produce sparse solutions. That is, all regression coefficients are generally non-zero with probability one. For better model interpretability, it may be preferable not only to shrink coefficients but also to enforce exact zero values for some of them. This enables simultaneous model estimation and variable selection. Such an approach is achieved by replacing the squared coefficient penalty with a penalty on the absolute values of the coefficients, i.e.,
\[
\text{pen}(\boldsymbol{\beta}) = \sum_{j=1}^{k} |\beta_j|.
\]
\noindent
This leads to the objective function for Lasso regression:
$$\text{LS}(\boldsymbol{\beta}) = (\mathbf{y} - \mathbf{X}\boldsymbol{\beta})^T(\mathbf{y} - \mathbf{X}\boldsymbol{\beta}) + \lambda \sum_{j=1}^{k} |\beta_j|,$$
\noindent
where $\lambda$ controls the strength of regularization.

LASSO (Lasso) stands for Least Absolute Shrinkage and Selection Operator~\cite{fahrmeir2022regression}. 
Ridge regression applies a quadratic penalty that heavily penalizes large coefficient values while exerting relatively little influence on those near zero . In contrast, the Lasso Regression uses an absolute value penalty that grows more slowly for large coefficients but increases more rapidly near zero. As a result, Lasso Regression tends to shrink smaller coefficients more aggressively toward zero, promoting sparsity, while having a reduced effect on larger coefficients.

\subsection{Support Vector Regression}

Support Vector Regression (SVR)~\cite{ma2003accurate} extends the principles of Support Vector Machines (SVM) to regression problems. Instead of minimizing the squared error as in traditional regression techniques, SVR aims to fit the best line within a specified threshold (called the $\epsilon$-tube) such that the predictions deviate from the true targets by at most $\epsilon$. 

Given a training dataset $T = \{(x_i, y_i)\}_{i=1}^{l}$, where $x_i \in \mathbb{R}^N$ and $y_i \in \mathbb{R}$, Support Vector Regression aims to construct a linear function of the form:
\begin{equation}
f(x) = W^T \phi(x) + b,
\end{equation}
where $W$ is a vector in the feature space $\mathcal{F}$ and $\phi(x)$ denotes a mapping from the input space to the feature space.

The parameters $W$ and $b$ are determined by solving the following convex optimization problem:
\begin{align}
\min_{W, b, \xi_i, \xi_i^*} \quad & \frac{1}{2} W^T W + C \sum_{i=1}^{l} (\xi_i + \xi_i^*) \\
\text{subject to} \quad & y_i - (W^T \phi(x_i) + b) \leq \epsilon + \xi_i \\
& (W^T \phi(x_i) + b) - y_i \leq \epsilon + \xi_i^* \\
& \xi_i, \xi_i^* \geq 0, \quad \text{for } i = 1, \dots, l.
\end{align}

Here, the loss function penalizes only those predictions where the absolute deviation from the target value exceeds a user-specified threshold $\epsilon$. The slack variables $\xi_i$ and $\xi_i^*$ represent the magnitude of positive and negative deviations respectively.

SVR is robust to outliers due to the $\epsilon$-insensitive loss function. The model complexity is controlled by $C$ and $\epsilon$, which must be carefully tuned. SVR can be extended with kernels (e.g., RBF, polynomial) to handle non-linear regression problems.

Unlike Ridge and Lasso, SVR does not impose a penalty directly on the coefficient magnitudes, but rather on the amount of error exceeding a certain threshold. This makes SVR particularly useful for applications where small deviations are acceptable, and focus is placed on limiting large errors.

\section{Proposed Algorithm}
\begin{figure*}[htbp]
    \centering
    \includegraphics[width=18cm]{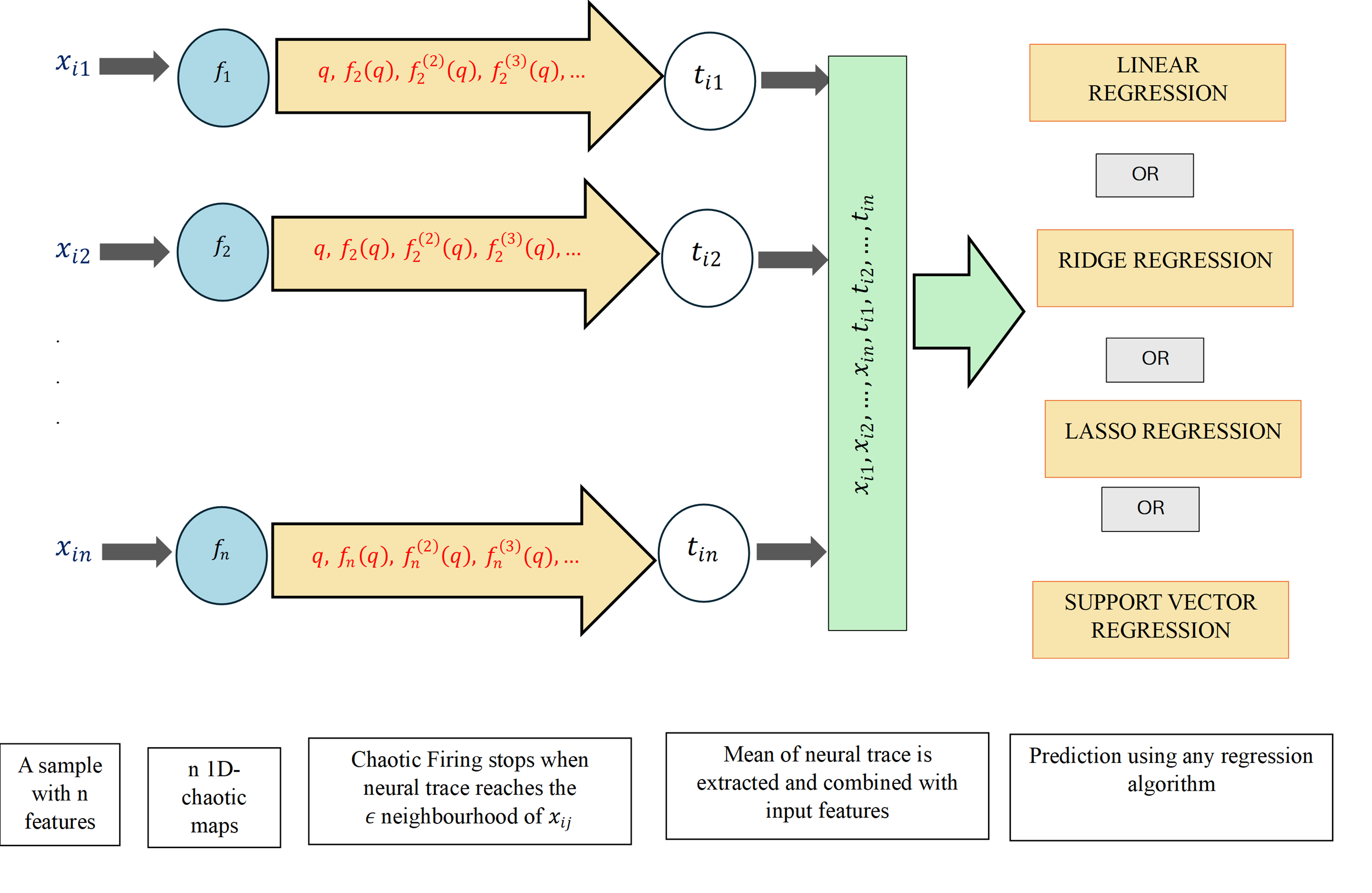}
    \caption{Augmented Chaotic Regression.}
    \label{fig:NL_Regression}
\end{figure*}

The proposed algorithm enhances traditional regression algorithms by incorporating both the original features and a chaotic feature—the arithmetic mean of the neural trace. This section outlines the key steps involved in the algorithm : 
\begin{itemize}
    \item \textit{Step 1} : Normalize the dataset 
    $$ \{(x_1^1, x_2^1, \dots, x_n^1), (x_1^2, x_2^2, \dots, x_n^2), \dots, (x_1^m, x_2^m, \dots, x_n^m)\}$$
    using min-max normalization. Here, there are $m$ samples with $n$ features. After normalisation let the dataset be $$ \{(z_1^1, z_2^1, \dots, z_n^1), (z_1^2, z_2^2, \dots, z_n^2), \dots, (z_1^m, z_2^m, \dots, z_n^m)\}$$
     \item \textit{Step 2 }: The input layer consists of $n$ 1D chaotic neurons . Here each neuron is a skew tent map with skew value $0.499$. Each neuron will start chaotic firing with an initial neural activity of $q$ units and the neural trace stops when it reaches the epsilon $\epsilon$ neighbourhood of $z_1^i, z_2^i, \dots, z_n^i $.
     \item \textit{Step 3 }: The only feature extracted from the chaotic neural trace corresponding to each input attribute $x_i^j$ is mean of neural trace, Tracemean (TM), denoted by $t_i^j$. 
Unlike from NL algorithms for classification, here both the original features and transformed features are used for regression.
Corresponding to the chaotic neural firing of the neuron in the input layer of NL (i.e., for each $z_i^j$), arithmetic mean of its neural trace is computed. Hence if the input data is of size $m\times n$ then the new feature extracted data is also of the size $m \times 2n$. Let the extracted feature set be :\\
$$\{(f^1_{1},f^1_{2},\ldots,f^1_{2n}),(f^2_1,f^2_2,\ldots,f^2_{2n}),\ldots,(f^m_1,f^m_2,\ldots,f^m_{2n}\}.$$\\
where $f_i^j=z_i^j$ for $j=1,2,\ldots n$ and $f_i^j=t_i^j$ for $j=n+1,n+2,\ldots 2n$.\\

\item \textit{Step 4 }: The extracted dataset, comprising input features and their corresponding mean of neural trace, can either be given as input to Linear Regression, Ridge Regression, Lasso Regression or Support Vector Regression. Accordingly, four distinct Neurochaos-based regression algorithms are proposed: 
\begin{itemize}
    \item Augmented Chaotic Linear Regression(ACLR)
     \item Augmented Chaotic Ridge Regression (ACRR)
    \item Augmented Chaotic Lasso Regression (ACLR)
    \item Augmented Chaotic Support Vector Regression (ACSVR)
\end{itemize}

\item \textit{Step 5} : Model will output a real valued predction for each sample.
\end{itemize}
The proposed algorithm is illustrated in figure \ref{fig:NL_Regression}.
\section{Results}
The performance of the proposed algorithms is analysed using two different ways : 
\begin{itemize}
    \item \textbf{Using ten different real life datasets} : The following datasets are used to analyse the performance of the algorithm: \textit{Diabetes Dataset, Bike Sharing Dataset, Fish, Life Expectancy, Concrete Strength, World Happiness Report, Body Fat, Auto MPG, Red Wine Quality, NASA Airfoil Self-Noise}. The description of datasets is included in Appendix \ref{regression_datasets}.
    
    \item \textbf{Using the generated dataset(synthetic dataset)}: The dataset generated is of the form $\{(x_i,y_i):y_i=mx_i+c+\epsilon,i=1,2,\ldots,n\}$. The value of $c$ is fixed as zero and $m$ is taken as $-2$ and $2$. The noise $\epsilon$ is randomly (Random state is fixed as 42 in our experiment) chosen from Normal distribution with mean zero and standard deviation varied among the values $5, 1$ and$0.1$. The value of $x_i$ is taken from the interval $(-10,10)$. Dataset of size \(10, 50, 100\) and \(1000\) are generated for comparison.  Thus we have considered $24$ generated datasets for comparing the performance of proposed four augumented chaotic regression algorithms.  
  
\begin{enumerate}
    \item $\{(x_i,y_i):y_i=2x_i+\epsilon,i=1,2,\ldots,10, \epsilon \in N(0,5)\}$
    \item $\{(x_i,y_i):y_i=-2x_i+\epsilon,i=1,2,\ldots,10, \epsilon \in N(0,5)\}$
     \item $\{(x_i,y_i):y_i=2x_i+\epsilon,i=1,2,\ldots,10, \epsilon \in N(0,1)\}$
    \item $\{(x_i,y_i):y_i=-2x_i+\epsilon,i=1,2,\ldots,10, \epsilon \in N(0,1)\}$
    \item $\{(x_i,y_i):y_i=2x_i+\epsilon,i=1,2,\ldots,10, \epsilon \in N(0,0.1)\}$
    \item $\{(x_i,y_i):y_i=-2x_i+\epsilon,i=1,2,\ldots,10, \epsilon \in N(0,0.1)\}$
    \item $\{(x_i,y_i):y_i=2x_i+\epsilon,i=1,2,\ldots,50, \epsilon \in N(0,5)\}$
    \item $\{(x_i,y_i):y_i=-2x_i+\epsilon,i=1,2,\ldots,50, \epsilon \in N(0,5)\}$
     \item $\{(x_i,y_i):y_i=2x_i+\epsilon,i=1,2,\ldots,50, \epsilon \in N(0,1)\}$
    \item $\{(x_i,y_i):y_i=-2x_i+\epsilon,i=1,2,\ldots,50, \epsilon \in N(0,1)\}$
    \item $\{(x_i,y_i):y_i=2x_i+\epsilon,i=1,2,\ldots,50, \epsilon \in N(0,0.1)\}$
    \item $\{(x_i,y_i):y_i=-2x_i+\epsilon,i=1,2,\ldots,50, \epsilon \in N(0,0.1)\}$
    \item $\{(x_i,y_i):y_i=2x_i+\epsilon,i=1,2,\ldots,100, \epsilon \in N(0,5)\}$
    \item $\{(x_i,y_i):y_i=-2x_i+\epsilon,i=1,2,\ldots,100, \epsilon \in N(0,5)\}$
     \item $\{(x_i,y_i):y_i=2x_i+\epsilon,i=1,2,\ldots,100, \epsilon \in N(0,1)\}$
    \item $\{(x_i,y_i):y_i=-2x_i+\epsilon,i=1,2,\ldots,100, \epsilon \in N(0,1)\}$
    \item $\{(x_i,y_i):y_i=2x_i+\epsilon,i=1,2,\ldots,100, \epsilon \in N(0,0.1)\}$
    \item $\{(x_i,y_i):y_i=-2x_i+\epsilon,i=1,2,\ldots,100, \epsilon \in N(0,0.1)\}$
    \item $\{(x_i,y_i):y_i=2x_i+\epsilon,i=1,2,\ldots,1000, \epsilon \in N(0,5)\}$
    \item $\{(x_i,y_i):y_i=-2x_i+\epsilon,i=1,2,\ldots,1000, \epsilon \in N(0,5)\}$
     \item $\{(x_i,y_i):y_i=2x_i+\epsilon,i=1,2,\ldots,1000, \epsilon \in N(0,1)\}$
    \item $\{(x_i,y_i):y_i=-2x_i+\epsilon,i=1,2,\ldots,1000, \epsilon \in N(0,1)\}$
    \item $\{(x_i,y_i):y_i=2x_i+\epsilon,i=1,2,\ldots,1000, \epsilon \in N(0,0.1)\}$
    \item $\{(x_i,y_i):y_i=-2x_i+\epsilon,i=1,2,\ldots,1000, \epsilon \in N(0,0.1)\}$
\end{enumerate}
Dataset of size $100$ for different values of $m$ and $\epsilon$ is shown in Figure \ref{fig:dataset100}.
\begin{figure}[htbp]
    \centering
    \begin{minipage}{0.48\textwidth}
        \centering
        \includegraphics[width=\textwidth]{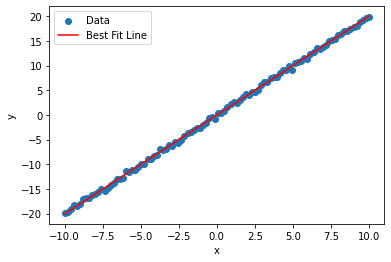}
        \caption*{$m=2, c=0, \epsilon \in N(0,0.1)$}
        \label{fig:fig1}
    \end{minipage}
    \hfill
    \begin{minipage}{0.48\textwidth}
        \centering
        \includegraphics[width=\textwidth]{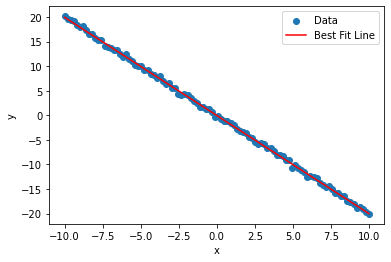}
        \caption*{$m=-2, c=0, \epsilon \in N(0,0.1)$}
        \label{fig:fig2}
    \end{minipage}
    
    \vspace{0.5cm} 
    
    \begin{minipage}{0.48\textwidth}
        \centering
        \includegraphics[width=\textwidth]{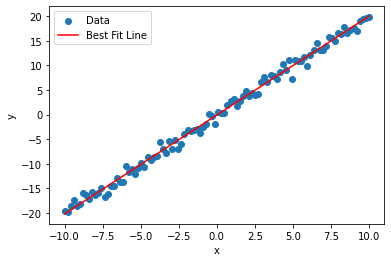}
        \caption*{$m=2, c=0, \epsilon \in N(0,1)$}
        \label{fig:fig3}
    \end{minipage}
    \hfill
    \begin{minipage}{0.48\textwidth}
        \centering
        \includegraphics[width=\textwidth]{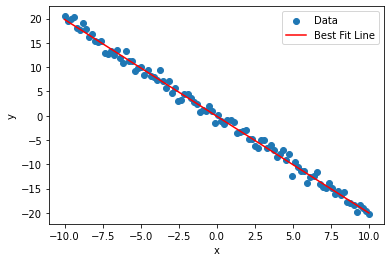}
        \caption*{$m=-2, c=0, \epsilon \in N(0,1)$}
        \label{fig:fig4}
    \end{minipage}

    \vspace{0.5cm}

    \begin{minipage}{0.48\textwidth}
        \centering
        \includegraphics[width=\textwidth]{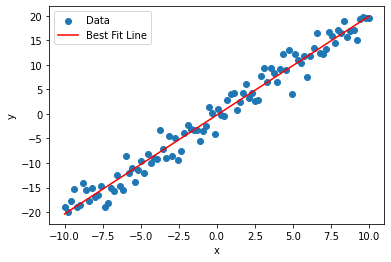}
        \caption*{$m=2, c=0, \epsilon \in N(0,5)$}
        \label{fig:fig5}
    \end{minipage}
    \hfill
    \begin{minipage}{0.48\textwidth}
        \centering
        \includegraphics[width=\textwidth]{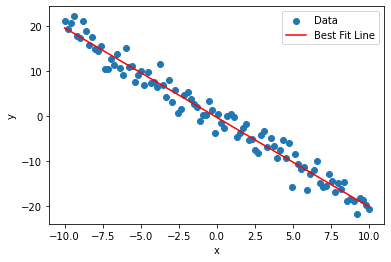}
        \caption*{$m=-2, c=0, \epsilon \in N(0,5)$}
        \label{fig:fig6}
    \end{minipage}

    \caption{Dataset of the form $\{(x_i,y_i):y_i=mx_i+c+\epsilon,i=1,2,\ldots,100\}$.}
    \label{fig:dataset100}
\end{figure}

Here the performance is compared with minimum mean square error(MMSE) corresponding to each dataset. MMSE is calculated by taking the difference between the predicted value and least square solution. Least square solution is computed using the pseudo-inverse method. The red line in  Figure \ref{fig:dataset100} represents the best fit line, the regression line that corresponds to MMSE.
\end{itemize}
\subsection{Hyperparameter Tuning}
Each dataset was initially split into training (80\%) and testing (20\%) sets. 
Training involves tuning of the hyperparameters associated with feature extraction - $q$ and epsilon (Noise around Stimulus) and the regression algorithm (eg: Regularisation Parameter in SVR) using five fold cross validation. The Initial neural activity, $q$ is tuned in the range $0.01$ to $0.99$ with a step value $0.01$ and epsilon (Noise around Stimulus) is tuned in the range $0.01$ to $0.45$ with a step value $0.01$.The regularisation parameter $C$ in SVR is tuned among the values $1, 10, 50, 100$. The regularisation parameter $\alpha$ in Lasso Regression and Ridge Regression is tuned among the values $0.1,1.0 \text{ and } 10.0$. The maximum number of iterations used in Lasso Regression is $10000$.

For real-world datasets, hyperparameter tuning was conducted with the objective of maximizing $R^2$ Score. The combination of hyperparameters that resulted in the highest $R^2$ Score was selected for model evaluation, as it was also observed to correspond to a low Mean Squared Error (MSE). Consequently, an independent tuning process based on MSE was deemed unnecessary. Following hyperparameter selection, model performance was assessed on a held-out test set comprising 20\% of the data, using evaluation metrics including the $R^2$ Score, MSE, and Mean Absolute Error (MAE).

In the case of synthetically generated datasets, hyperparameters were optimized by minimizing the MSE. Accordingly, model performance on these datasets was evaluated primarily using the MSE metric.

The following section will present the value of hyperparameters and performance metrics for different datasets using different versions of augmented chaotic regression algorithms.

\subsection{Augmented Chaotic Linear Regression}
This section outlines the performance of the proposed algorithm, Augmented Chaotic Linear Regression, using both real world and generated datasets. This algorithm requires the tuning of only two hyperparameters $q$ and epsilon (Noise around Stimulus). The table \ref{tab:ACLR_real} presents the value of hyperparameters and the performance metrics $R^2$ Score, MSE and MAE for real world datasets. Tables \ref{tab: ACLR_10}, \ref{tab: ACLR_50}, \ref{tab: ACLR_100}, \ref{tab: ACLR_1000} correseponds to the dataset of size $10$, $50$, $100$ and $1000$ respectively

\begin{table}[h]
    \centering
     \caption{\textbf{Hyperparameters and Performance metrics of Augmented Chaotic LR on real life datasets.}}
   
    \begin{tabular}{clccccccll}
        \toprule
\textbf{Sl No} & \textbf{Datasets} & \textbf{q} & \textbf{epsilon} & \textbf{Training R\textsuperscript{2} Score} & \textbf{Testing R\textsuperscript{2} Score} & \textbf{MSE} & \textbf{MAE} \\
        \midrule
1 & Diabetes Dataset & 0.11 & 0.21 & 0.505 & 0.271 & 3859.203 & 49.444 \\
2 & Bike Sharing Dataset & 0.01 & 0.01 & 0.999 & 0.962 & 152936.267 & 372.184 \\
3 & Fish & 0.03 & 0.44 & 0.952 & 0.875 & 17767.677 & 107.112 \\
4 & Life Expectancy & 0.18 & 0.11 & 0.887 & 0.845 & 10.994 & 2.398 \\
5 & Concrete & 0.65 & 0.08 & 0.802 & 0.776 & 64.148 & 5.97 \\
6 & Happiness & 0.08 & 0.03 & 0.804 & 0.474 & 0.547 & 0.543 \\
7 & bodyfat & 0.05 & 0.05 & 0.968 & 0.698 & 14.011 & 3.063 \\
8 & autompg & 0.48 & 0.07 & 0.762 & 0.658 & 17.453 & 3.174 \\
9 & Wine quality & 0.84 & 0.26 & 0.682 & 0.648 & 0.438 & 0.493 \\
10 & Airfoil noise & 0.19 & 0.03 & 0.520 & 0.548 & 22.646 & 3.708 \\
\bottomrule
\end{tabular}
   
\label{tab:ACLR_real}
\end{table}

\begin{table}[htbp]
\centering
\caption{\textbf{Hyperparameters and Performance metrics of Augmented Chaotic LR on dataset of the form $\{(x_i,y_i):y_i=mx_i+c+\epsilon,i=1,2,\ldots,n\}$ with sample size $n=10$}.}

\begin{tabular}{lcccccc}
\toprule
\textbf{Parameters to generate dataset} & \textbf{q} & \textbf{epsilon} & \textbf{Training MSE} & \textbf{Testing MSE} \\
\midrule
$\text{Size}=10$ , $m=2$ , $c=0$ , $\epsilon \in N(0,5)$        & 0.01 & 0.08 & 1.236 & 23.891 \\
$\text{Size}=10$ , $m=-2$ , $c=0$ , $\epsilon \in N(0,5)$       & 0.01 & 0.08 & 1.236 & 22.142 \\
$\text{Size}=10$ , $m=2$ , $c=0$ , $\epsilon \in N(0,1)$        & 0.01 & 0.08 & 0.247 & 20.797 \\
$\text{Size}=10$ , $m=-2$ , $c=0$ , $\epsilon \in N(0,1)$       & 0.01 & 0.08 & 0.247 & 20.015 \\
$\text{Size}=10$ , $m=2$ , $c=0$ , $\epsilon \in N(0,0.1)$      & 0.01 & 0.08 & 0.025 & 19.940 \\
$\text{Size}=10$ , $m=-2$ , $c=0$ , $\epsilon \in N(0,0.1)$     & 0.01 & 0.08 & 0.025 & 19.694 \\
\bottomrule
\end{tabular}

\label{tab: ACLR_10}
\end{table}

\begin{table}[htbp]
\centering
\caption{\textbf{Hyperparameters and Performance metrics of Augmented Chaotic LR on dataset of the form $\{(x_i,y_i):y_i=mx_i+c+\epsilon,i=1,2,\ldots,n\}$ with sample size $n=50$.}}

\begin{tabular}{lcccccc}
\toprule
Parameters to generate dataset & q & epsilon & Training MSE & Testing MSE \\
\midrule
$\text{Size}=50$ , $m=2$ , $c=0$ , $\epsilon \in N(0,5)$        & 0.58 & 0.05 & 3.942 & 31.923 \\
$\text{Size}=50$ , $m=-2$ , $c=0$ , $\epsilon \in N(0,5)$       & 0.80 & 0.01 & 4.079 & 64.763 \\
$\text{Size}=50$ , $m=2$ , $c=0$ , $\epsilon \in N(0,1)$        & 0.80 & 0.01 & 0.815 & 33.778 \\
$\text{Size}=50$ , $m=-2$ , $c=0$ , $\epsilon \in N(0,1)$       & 0.80 & 0.01 & 0.815 & 50.293 \\
$\text{Size}=50$ , $m=2$ , $c=0$ , $\epsilon \in N(0,0.1)$      & 0.80 & 0.01 & 0.081 & 38.465 \\
$\text{Size}=50$ , $m=-2$ , $c=0$ , $\epsilon \in N(0,0.1)$     & 0.80 & 0.01 & 0.081 & 43.687 \\
\bottomrule
\end{tabular}

\label{tab: ACLR_50}
\end{table}

\begin{table}[htbp]
\centering
\caption{\textbf{Hyperparameters and Performance metrics of Augmented Chaotic LR on dataset of the form $\{(x_i,y_i):y_i=mx_i+c+\epsilon,i=1,2,\ldots,n\}$ with sample size $n=100$.}}

\begin{tabular}{lcccccc}
\toprule
Parameters to generate dataset  & q & epsilon & Training MSE & Testing MSE \\
\midrule
$\text{Size}=100$ , $m=2$ , $c=0$ , $\epsilon \in N(0,5)$        & 0.36 & 0.02 & 4.161 & 7.181 \\
$\text{Size}=100$ , $m=-2$ , $c=0$ , $\epsilon \in N(0,5)$       & 0.02 & 0.01 & 4.067 & 9.020 \\
$\text{Size}=100$ , $m=2$ , $c=0$ , $\epsilon \in N(0,1)$        & 0.02 & 0.01 & 0.873 & 5.158 \\
$\text{Size}=100$ , $m=-2$ , $c=0$ , $\epsilon \in N(0,1)$       & 0.01 & 0.02 & 0.802 & 6.295 \\
$\text{Size}=100$ , $m=2$ , $c=0$ , $\epsilon \in N(0,0.1)$      & 0.77 & 0.01 & 0.089 & 44.400 \\
$\text{Size}=100$ , $m=-2$ , $c=0$ , $\epsilon \in N(0,0.1)$     & 0.01 & 0.02 & 0.082 & 5.289 \\
\bottomrule
\end{tabular}

\label{tab: ACLR_100}
\end{table}

\begin{table}[htbp]
\centering
\caption{\textbf{Hyperparameters and Performance metrics of Augmented Chaotic LR on dataset of the form $\{(x_i,y_i):y_i=mx_i+c+\epsilon,i=1,2,\ldots,n\}$ with sample size $n=1000$.}}

\begin{tabular}{lcccccc}
\toprule
Parameters to generate dataset  & q & epsilon & Training MSE & Testing MSE \\
\midrule
$\text{Size}=1000$ , $m=2$ , $c=0$ , $\epsilon \in N(0,5)$        & 0.02 & 0.01 & 4.795 & 4.795 \\
$\text{Size}=1000$ , $m=-2$ , $c=0$ , $\epsilon \in N(0,5)$       & 0.01 & 0.01 & 4.835 & 4.847 \\
$\text{Size}=1000$ , $m=2$ , $c=0$ , $\epsilon \in N(0,1)$        & 0.01 & 0.01 & 0.959 & 0.985 \\
$\text{Size}=1000$ , $m=-2$ , $c=0$ , $\epsilon \in N(0,1)$       & 0.01 & 0.01 & 0.974 & 1.018 \\
$\text{Size}=1000$ , $m=2$ , $c=0$ , $\epsilon \in N(0,0.1)$      & 0.01 & 0.01 & 0.098 & 0.139 \\
$\text{Size}=1000$ , $m=-2$ , $c=0$ , $\epsilon \in N(0,0.1)$     & 0.01 & 0.01 & 0.103 & 0.150 \\

\bottomrule
\end{tabular}

\label{tab: ACLR_1000}
\end{table}

\subsection{Augmented Chaotic Ridge Regression}
This section outlines the performance of the proposed algorithm Augmented Chaotic Ridge Regression using both real world and generated datasets. This algorithm requires the tuning three hyperparameters $q$, epsilon (Noise around Stimulus) and $\alpha$. The table \ref{tab:ACRR_real} presents the value of hyperparameters and the performance metrics $R^2$ Score, MSE and MAE for real world datasets. Tables \ref{tab: ACRR_10}, \ref{tab: ACRR_50}, \ref{tab: ACRR_100}, \ref{tab: ACRR_1000} correseponds to the dataset of size $10$, $50$, $100$ and $1000$ respectively.

\begin{table}[htbp]
    \centering
     \caption{\textbf{Hyperparameters and Performance metrics of Augmented Chaotic RR on real life datasets.}}
   
    \begin{tabular}{clccccccll}
        \toprule
\textbf{Sl no} & \textbf{Datasets} & \textbf{alpha} & \textbf{q} & \textbf{epsilon} & \textbf{Training R\textsuperscript{2} Score} & \textbf{Testing R\textsuperscript{2} Score} & \textbf{MSE} & \textbf{MAE} \\
\midrule
1 & Diabetes Dataset & 1 & 0.57 & 0.069 & 0.509 & 0.442 & 2953.824 & 42.774 \\
2 & Bike Sharing Dataset & 0.1 & 0.01 & 0.01 & 0.999 & 0.962 & 151973.922 & 370.723 \\
3 & Fish & 0.1 & 0.79 & 0.15 & 0.953 & 0.954 & 6516.313 & 67.618 \\
4 & Life Expectancy & 0.1 & 0.18 & 0.11 & 0.885 & 0.846 & 10.925 & 2.401 \\
5 & Concrete & 0.1 & 0.069 & 0.05 & 0.794 & 0.788 & 60.61 & 6.075 \\
6 & Happiness & 0.1 & 0.03 & 0.06 & 0.809 & 0.509 & 0.51 & 0.531 \\
7 & Bodyfat & 0.1 & 0.05 & 0.05 & 0.969 & 0.749 & 2.78 & 11.685 \\
8 & AutoMPG & 0.1 & 0.48 & 0.069 & 0.763 & 0.658 & 17.441 & 3.187 \\
9 & Wine Quality & 0.1 & 0.84 & 0.26 & 0.682 & 0.649 & 0.437 & 0.494 \\
10 & Airfoil Noise & 0.1 & 0.19 & 0.03 & 0.52 & 0.548 & 22.622 & 3.707 \\
        \bottomrule
    \end{tabular}
   
    \label{tab:ACRR_real}
\end{table}

\begin{table}[htbp]
\centering
\caption{\textbf{Hyperparameters and Performance metrics of Augmented Chaotic RR on dataset of the form $\{(x_i,y_i):y_i=mx_i+c+\epsilon,i=1,2,\ldots,n\}$ with sample size $n=10$.}}

\begin{tabular}{lcccccc}
\toprule
\textbf{Parameters to generate dataset} & \textbf{alpha} & \textbf{q} & \textbf{epsilon} & \textbf{Training MSE} & \textbf{Testing MSE} \\
\midrule
Size = 10, $m=2$, $c=0$, $\epsilon \in N(0,5)$  & 0.1 & 0.08 & 0.13 & 10.127 & 17.227 \\
Size = 10, $m=-2$, $c=0$, $\epsilon \in N(0,5)$  & 0.1 & 0.01 & 0.069 & 8.596 & 7.844 \\
Size = 10, $m=2$, $c=0$, $\epsilon \in N(0,1)$  & 0.1 & 0.08 & 0.13 & 7.652 & 10.590 \\
Size = 10, $m=-2$, $c=0$, $\epsilon \in N(0,1)$  & 0.1 & 0.08 & 0.13 & 7.188 & 5.732 \\
Size = 10, $m=2$, $c=0$, $\epsilon \in N(0,0.1)$  & 0.1 & 0.08 & 0.13 & 7.001 & 8.111 \\
Size = 10, $m=-2$, $c=0$, $\epsilon \in N(0,0.1)$  & 0.1 & 0.08 & 0.13 & 6.854 & 6.575 \\

\bottomrule
\end{tabular}

\label{tab: ACRR_10}
\end{table}

\begin{table}[htbp]
\centering
\caption{\textbf{Hyperparameters and Performance metrics of Augmented Chaotic RR on dataset of the form $\{(x_i,y_i):y_i=mx_i+c+\epsilon,i=1,2,\ldots,n\}$ with sample size $n=50$.}}

\begin{tabular}{lcccccc}
\toprule
\textbf{Parameters to generate dataset} & \textbf{alpha} & \textbf{q} & \textbf{epsilon} & \textbf{Training MSE} & \textbf{Testing MSE} \\
\midrule
Size = 50, $m=2$, $c=0$, $\epsilon \in N(0,5)$  & 0.1 & 0.32 & 0.03 & 4.455 & 30.842 \\
Size = 50, $m=-2$, $c=0$, $\epsilon \in N(0,5)$  & 0.1 & 0.06 & 0.01 & 4.364 & 58.846 \\
Size = 50, $m=2$, $c=0$, $\epsilon \in N(0,1)$  & 0.1 & 0.65 & 0.01 & 1.067 & 32.829 \\
Size = 50, $m=-2$, $c=0$, $\epsilon \in N(0,1)$  & 0.1 & 0.8 & 0.01 & 1.019 & 47.270 \\
Size = 50, $m=2$, $c=0$, $\epsilon \in N(0,0.1)$  & 0.1 & 0.26 & 0.01 & 0.263 & 35.873 \\
Size = 50, $m=-2$, $c=0$, $\epsilon \in N(0,0.1)$  & 0.1 & 0.8 & 0.01 & 0.254 & 41.095 \\

\bottomrule
\end{tabular}

\label{tab: ACRR_50}
\end{table}

\begin{table}[htbp]
\centering
\caption{\textbf{Hyperparameters and Performance metrics of Augmented Chaotic RR on dataset of the form $\{(x_i,y_i):y_i=mx_i+c+\epsilon,i=1,2,\ldots,n\}$ with sample size $n=100$.}}

\begin{tabular}{lcccccc}
\toprule
\textbf{Parameters to generate dataset} & \textbf{alpha} & \textbf{q} & \textbf{epsilon} & \textbf{Training MSE} & \textbf{Testing MSE} \\
\midrule
Size = 100, $m=2$, $c=0$, $\epsilon \in N(0,5)$  & 0.1 & 0.02 & 0.01 & 4.253 & 6.633 \\
Size = 100, $m=-2$, $c=0$, $\epsilon \in N(0,5)$  & 0.1 & 0.02 & 0.01 & 4.104 & 8.763 \\
Size = 100, $m=2$, $c=0$, $\epsilon \in N(0,1)$  & 0.1 & 0.02 & 0.01 & 0.915 & 4.882 \\
Size = 100, $m=-2$, $c=0$, $\epsilon \in N(0,1)$  & 0.1 & 0.02 & 0.01 & 0.849 & 5.835 \\
Size = 100, $m=2$, $c=0$, $\epsilon \in N(0,0.1)$  & 0.1 & 0.3 & 0.01 & 0.133 & 4.687 \\
Size = 100, $m=-2$, $c=0$, $\epsilon \in N(0,0.1)$  & 0.1 & 0.02 & 0.01 & 0.130 & 4.983 \\
\bottomrule
\end{tabular}

\label{tab: ACRR_100}
\end{table}

\begin{table}[htbp]
\centering
\caption{\textbf{Hyperparameters and Performance metrics of Augmented Chaotic RR on dataset of the form $\{(x_i,y_i):y_i=mx_i+c+\epsilon,i=1,2,\ldots,n\}$ with sample size $n=1000$.}}

\begin{tabular}{lcccccc}
\toprule
\textbf{Parameters to generate dataset} & \textbf{alpha} & \textbf{q} & \textbf{epsilon} & \textbf{Training MSE} & \textbf{Testing MSE} \\
\midrule
Size = 1000, $m=2$, $c=0$, $\epsilon \in N(0,5)$  & 0.1 & 0.02 & 0.01 & 4.795 & 4.792 \\
Size = 1000, $m=-2$, $c=0$, $\epsilon \in N(0,5)$  & 0.1 & 0.01 & 0.01 & 4.835 & 4.841 \\
Size = 1000, $m=2$, $c=0$, $\epsilon \in N(0,1)$  & 0.1 & 0.01 & 0.01 & 0.959 & 0.980 \\
Size = 1000, $m=-2$, $c=0$, $\epsilon \in N(0,1)$  & 0.1 & 0.01 & 0.01 & 0.975 & 1.013 \\
Size = 1000, $m=2$, $c=0$, $\epsilon \in N(0,0.1)$  & 0.1 & 0.01 & 0.01 & 0.098 & 0.135 \\
Size = 1000, $m=-2$, $c=0$, $\epsilon \in N(0,0.1)$  & 0.1 & 0.01 & 0.01 & 0.103 & 0.145 \\

\bottomrule
\end{tabular}

\label{tab: ACRR_1000}
\end{table}

\subsection{Augmented Chaotic Lasso Regression}
This section outlines the performance of the proposed algorithm Augmented Chaotic Lasso Regression using both real world and generated datasets. This algorithm requires the tuning three hyperparameters $q$, epsilon (Noise around Stimulus) and $\alpha$. The table \ref{tab:ACLS_real} presents the value of hyperparameters and the performance metrics $R^2$ Score, MSE and MAE for real world datasets. Tables \ref{tab: ACLS_10}, \ref{tab: ACLS_50}, \ref{tab: ACLS_100}, \ref{tab: ACLS_1000} correseponds to the dataset of size $10$, $50$, $100$ and $1000$ respectively

\begin{table}[htbp]
    \centering
     \caption{\textbf{Hyperparameters and Performance metrics of Augmented Chaotic LS on real life datasets.}}
  
    \begin{tabular}{clcccccll}
        \toprule
\textbf{Sl No} & \textbf{Dataset} & \textbf{alpha} & \textbf{q} & \textbf{epsilon} & \textbf{Training $R^2$ Score} & \textbf{Testing $R^2$ Score} & \textbf{MSE} & \textbf{MAE} \\
\midrule
1 & Diabetes & 0.1 & 0.57 & 0.08 & 0.506 & 0.434 & 2998.53 & 43.137 \\
2 & Bike Sharing & 0.1 & 0.01 & 0.01 & 0.999 & 0.961 & 153835.585 & 373.370 \\
3 & Fish & 0.1 & 0.03 & 0.44 & 0.955 & 0.883 & 16597.001 & 104.213 \\
4 & Life Expectancy & 0.1 & 0.18 & 0.11 & 0.852 & 0.829 & 12.111 & 2.463 \\
5 & Concrete & 0.1 & 0.069 & 0.05 & 0.781 & 0.787 & 60.935 & 5.873 \\
6 & Happiness & 0.1 & 0.35 & 0.26 & 0.677 & 0.577 & 0.440 & 0.539 \\
7 & Body Fat & 0.1 & 0.18 & 0.01 & 0.966 & 0.748 & 11.681 & 2.576 \\
8 & Auto MPG & 0.1 & 0.52 & 0.09 & 0.754 & 0.678 & 16.403 & 3.171 \\
9 & Wine Quality\textsuperscript{*} & 0.0001 & 0.84 & 0.26 & 0.682 & 0.651 & 0.434 & 0.491 \\
10 & Airfoil Noise & 0.1 & 0.98 & 0.02 & 0.481 & 0.514 & 24.305 & 3.931 \\

        \bottomrule
    \end{tabular}
    \vspace{1mm}
\begin{flushleft}
\textsuperscript{*}\textit{Note:} Unlike the other datasets, the Wine quality dataset required tuning of $\alpha$ from the smaller values 0.0001, 0.001, and 0.01. Larger values such as 0.1, 1, and 10 resulted in an $R^2$ Score of 0 for standalone Lasso Regression.
\end{flushleft}
   
    \label{tab:ACLS_real}

\end{table}

\begin{table}[htbp]
\centering
\caption{\textbf{Hyperparameters and Performance metrics of Augmented Chaotic LS on dataset of the form $\{(x_i,y_i):y_i=mx_i+c+\epsilon,i=1,2,\ldots,n\}$ with sample size $n=10$.}}

\begin{tabular}{lcccccc}
\toprule
\textbf{Dataset Parameters} & \textbf{alpha} & \textbf{q} & \textbf{epsilon} & \textbf{Training MSE} & \textbf{Testing MSE} \\
\midrule
Size = 10, $m=2$, $c=0$, $\epsilon \in N(0,5)$  & 0.1 & 0.38 & 0.08 & 3.032 & 11.342 \\
Size = 10, $m=-2$, $c=0$, $\epsilon \in N(0,5)$  & 0.1 & 0.02 & 0.069 & 1.715 & 16.637 \\
Size = 10, $m=2$, $c=0$, $\epsilon \in N(0,1)$  & 0.1 & 0.01 & 0.01 & 0.921 & 17.670 \\
Size = 10, $m=-2$, $c=0$, $\epsilon \in N(0,1)$  & 0.1 & 0.02 & 0.069 & 0.791 & 14.581 \\
Size = 10, $m=2$, $c=0$, $\epsilon \in N(0,0.1)$  & 0.1 & 0.01 & 0.01 & 0.418 & 15.711 \\
Size = 10, $m=-2$, $c=0$, $\epsilon \in N(0,0.1)$  & 0.1 & 0.01 & 0.01 & 0.412 & 14.672 \\

\bottomrule
\end{tabular}

\label{tab: ACLS_10}
\end{table}

\begin{table}[htbp]
\centering
\caption{\textbf{Hyperparameters and Performance metrics of Augmented Chaotic LS on dataset of the form $\{(x_i,y_i):y_i=mx_i+c+\epsilon,i=1,2,\ldots,n\}$ with sample size $n=50$.}}

\begin{tabular}{lcccccc}
\toprule
\textbf{Dataset Parameters} & \textbf{alpha} & \textbf{q} & \textbf{epsilon} & \textbf{Training MSE} & \textbf{Testing MSE} \\
\midrule
Size = 50, $m=2$, $c=0$, $\epsilon \in N(0,5)$  & 0.1 & 0.01 & 0.02 & 4.812 & 25.765 \\
Size = 50, $m=-2$, $c=0$, $\epsilon \in N(0,5)$  & 0.1 & 0.05 & 0.02 & 4.469 & 57.896 \\
Size = 50, $m=2$, $c=0$, $\epsilon \in N(0,1)$  & 0.1 & 0.01 & 0.02 & 1.067 & 31.385 \\
Size = 50, $m=-2$, $c=0$, $\epsilon \in N(0,1)$  & 0.1 & 0.06 & 0.01 & 1.004 & 45.760 \\
Size = 50, $m=2$, $c=0$, $\epsilon \in N(0,0.1)$  & 0.1 & 0.01 & 0.01 & 0.220 & 36.148 \\
Size = 50, $m=-2$, $c=0$, $\epsilon \in N(0,0.1)$  & 0.1 & 0.01 & 0.01 & 0.204 & 41.018 \\

\bottomrule
\end{tabular}

\label{tab: ACLS_50}
\end{table}

\begin{table}[htbp]
\centering
\caption{\textbf{Hyperparameters and Performance metrics of Augmented Chaotic LS on dataset of the form $\{(x_i,y_i):y_i=mx_i+c+\epsilon,i=1,2,\ldots,n\}$ with sample size $n=100$.}}

\begin{tabular}{lcccccc}
\toprule
\textbf{Dataset Parameters} & \textbf{alpha} & \textbf{q} & \textbf{epsilon} & \textbf{Training MSE} & \textbf{Testing MSE} \\
\midrule
Size = 100, $m=2$, $c=0$, $\epsilon \in N(0,5)$  & 0.1 & 0.13 & 0.01 & 4.254 & 8.102 \\
Size = 100, $m=-2$, $c=0$, $\epsilon \in N(0,5)$  & 0.1 & 0.01 & 0.06 & 4.297 & 8.540 \\
Size = 100, $m=2$, $c=0$, $\epsilon \in N(0,1)$  & 0.1 & 0.3 & 0.01 & 1.006 & 4.774 \\
Size = 100, $m=-2$, $c=0$, $\epsilon \in N(0,1)$  & 0.1 & 0.01 & 0.06 & 0.957 & 5.602 \\
Size = 100, $m=2$, $c=0$, $\epsilon \in N(0,0.1)$  & 0.1 & 0.3 & 0.01 & 0.207 & 4.451 \\
Size = 100, $m=-2$, $c=0$, $\epsilon \in N(0,0.1)$  & 0.1 & 0.01 & 0.02 & 0.204 & 4.700 \\
\bottomrule
\end{tabular}

\label{tab: ACLS_100}
\end{table}

\begin{table}[htbp]
\centering
\caption{\textbf{Hyperparameters and Performance metrics of Augmented Chaotic LS on dataset of the form $\{(x_i,y_i):y_i=mx_i+c+\epsilon,i=1,2,\ldots,n\}$ with sample size $n=1000$.}}

\begin{tabular}{lcccccc}
\toprule
\textbf{Dataset Parameters} & \textbf{alpha} & \textbf{q} & \textbf{epsilon} & \textbf{Training MSE} & \textbf{Testing MSE} \\
\midrule
Size = 1000, $m=2$, $c=0$, $\epsilon \in N(0,5)$  & 0.1 & 0.01 & 0.03 & 4.903 & 4.755 \\
Size = 1000, $m=-2$, $c=0$, $\epsilon \in N(0,5)$  & 0.1 & 0.01 & 0.01 & 4.950 & 4.888 \\
Size = 1000, $m=2$, $c=0$, $\epsilon \in N(0,1)$  & 0.1 & 0.01 & 0.01 & 1.075 & 1.012 \\
Size = 1000, $m=-2$, $c=0$, $\epsilon \in N(0,1)$  & 0.1 & 0.01 & 0.01 & 1.094 & 1.059 \\
Size = 1000, $m=2$, $c=0$, $\epsilon \in N(0,0.1)$  & 0.1 & 0.01 & 0.01 & 0.217 & 0.173 \\
Size = 1000, $m=-2$, $c=0$, $\epsilon \in N(0,0.1)$  & 0.1 & 0.01 & 0.01 & 0.223 & 0.188 \\

\bottomrule
\end{tabular}

\label{tab: ACLS_1000}
\end{table}

\subsection{Augmented Chaotic SVR}
This section outlines the performance of the proposed algorithm Augmented chaotic SVR Regression using both real world and generated datasets. This algorithm requires the tuning three hyperparameters $q$, epsilon (Noise around Stimulus) and $C$. For real world datasets, the kernel used is Radial Basis Function (RBF). But for generated data, the kernel used linear function. The table \ref{tab:ACSVR_real} presents the value of hyperparameters and the performance metrics $R^2$ Score, MSE and MAE for real world datasets. Tables \ref{tab: ACSVR_10}, \ref{tab: ACSVR_50}, \ref{tab: ACSVR_100}, \ref{tab: ACSVR_1000} correseponds to the dataset of size $10$, $50$, $100$ and $1000$ respectively.

\begin{table}[htbp]
    \centering
     \caption{\textbf{Hyperparameters and Performance metrics of Augmented Chaotic SVR on real life datasets.}}

    \begin{tabular}{clccccccll}
        \toprule
\textbf{Sl. No} & \textbf{Dataset} & \textbf{q} & \textbf{epsilon} & \textbf{c} & \textbf{Training $R^2$ Score} & \textbf{Testing $R^2$ score} & \textbf{MSE} & \textbf{MAE} \\
\midrule
1 & Diabetes Dataset & 0.05 & 0.099 & 50 & 0.478 & 0.471 & 2801.659 & 40.605 \\
2 & Bike Sharing Dataset & 0.03 & 0.12 & 100 & 0.885 & 0.900 & 400155.243 & 455.279 \\
3 & Fish & 0.069 & 0.29 & 100 & 0.902 & 0.970 & 4258.662 & 49.033 \\
4 & Life Expectancy & 0.12 & 0.05 & 100 & 0.931 & 0.914 & 6.132 & 1.661 \\
5 & Concrete & 0.5 & 0.01 & 100 & 0.910 & 0.854 & 41.860 & 4.690 \\
6 & Happiness & 0.04 & 0.34 & 50 & 0.845 & 0.620 & 0.395 & 0.484 \\
7 & Bodyfat & 0.03 & 0.12 & 50 & 0.975 & 0.829 & 7.960 & 2.133 \\
8 & Auto MPG & 0.51 & 0.01 & 50 & 0.788 & 0.653 & 17.680 & 3.029 \\
9 & Wine Quality & 0.89 & 0.01 & 10 & 0.754 & 0.718 & 0.350 & 0.438 \\
10 & Airfoil Noise & 0.75 & 0.02 & 100 & 0.835 & 0.860 & 7.011 & 2.046 \\

        \bottomrule
    \end{tabular}
    \label{tab:ACSVR_real}
\end{table}

\begin{table}[htbp]
\centering
\caption{\textbf{Hyperparameters and Performance metrics of Augmented Chaotic SVR on dataset of the form $\{(x_i,y_i):y_i=mx_i+c+\epsilon,i=1,2,\ldots,n\}$ with sample size $n=10$.}}

\begin{tabular}{lcccccc}
\toprule
\textbf{Dataset Parameters} & \textbf{alpha} & \textbf{q} & \textbf{epsilon} & \textbf{Training MSE} & \textbf{Testing MSE} \\
\midrule
Size = 10, $m=2$, $c=0$, $\epsilon \in N(0,5)$      & 0.1 & 0.38 & 0.099 & 1.918  & 8.694  \\
Size = 10, $m=-2$, $c=0$, $\epsilon \in N(0,5)$     & 0.1 & 0.08 & 0.060 & 1.093  & 16.105 \\
Size = 10, $m=2$, $c=0$, $\epsilon \in N(0,1)$      & 0.1 & 0.01 & 0.069 & 0.374  & 22.124 \\
Size = 10, $m=-2$, $c=0$, $\epsilon \in N(0,1)$     & 0.1 & 0.08 & 0.060 & 0.191  & 16.693 \\
Size = 10, $m=2$, $c=0$, $\epsilon \in N(0,0.1)$    & 0.1 & 0.65 & 0.010 & 0.022  & 20.315 \\
Size = 10, $m=-2$, $c=0$, $\epsilon \in N(0,0.1)$   & 0.1 & 0.06 & 0.080 & 0.0198 & 19.152 \\

\bottomrule
\end{tabular}

\label{tab: ACSVR_10}
\end{table}

\begin{table}[htbp]
\centering
\caption{\textbf{Hyperparameters and Performance metrics of Augmented Chaotic SVR on dataset of the form $\{(x_i,y_i):y_i=mx_i+c+\epsilon,i=1,2,\ldots,n\}$ with sample size $n=50$.}}

\begin{tabular}{lcccccc}
\toprule
\textbf{Dataset Parameters} & \textbf{alpha} & \textbf{q} & \textbf{epsilon} & \textbf{Training MSE} & \textbf{Testing MSE} \\
\midrule
Size = 50, $m=2$, $c=0$, $\epsilon \in N(0,5)$      & 0.1 & 0.8  & 0.01 & 4.164  & 27.708  \\
Size = 50, $m=-2$, $c=0$, $\epsilon \in N(0,5)$     & 0.1 & 0.06 & 0.01 & 4.198  & 57.440  \\
Size = 50, $m=2$, $c=0$, $\epsilon \in N(0,1)$      & 0.1 & 0.35 & 0.01 & 0.879  & 56.960  \\
Size = 50, $m=-2$, $c=0$, $\epsilon \in N(0,1)$     & 0.1 & 0.8  & 0.01 & 0.845  & 50.239  \\
Size = 50, $m=2$, $c=0$, $\epsilon \in N(0,0.1)$    & 0.1 & 0.8  & 0.01 & 0.0857 & 38.241  \\
Size = 50, $m=-2$, $c=0$, $\epsilon \in N(0,0.1)$   & 0.1 & 0.8  & 0.01 & 0.0840 & 43.828  \\

\bottomrule
\end{tabular}

\label{tab: ACSVR_50}
\end{table}

\begin{table}[htbp]
\centering
\caption{\textbf{Hyperparameters and Performance metrics of Augmented Chaotic SVR on dataset of the form $\{(x_i,y_i):y_i=mx_i+c+\epsilon,i=1,2,\ldots,n\}$ with sample size $n=100$.}}

\begin{tabular}{lcccccc}
\toprule
\textbf{Dataset Parameters} & \textbf{alpha} & \textbf{q} & \textbf{epsilon} & \textbf{Training MSE} & \textbf{Testing MSE} \\
\midrule
Size = 100, $m=2$, $c=0$, $\epsilon \in N(0,5)$      & 0.1 & 0.36 & 0.02 & 4.135 & 8.101 \\
Size = 100, $m=-2$, $c=0$, $\epsilon \in N(0,5)$     & 0.1 & 0.02 & 0.01 & 3.989 & 8.795 \\
Size = 100, $m=2$, $c=0$, $\epsilon \in N(0,1)$      & 0.1 & 0.04 & 0.01 & 0.874 & 5.137 \\
Size = 100, $m=-2$, $c=0$, $\epsilon \in N(0,1)$     & 0.1 & 0.02 & 0.01 & 0.798 & 5.820 \\
Size = 100, $m=2$, $c=0$, $\epsilon \in N(0,0.1)$    & 0.1 & 0.77 & 0.01 & 0.089 & 43.896 \\
Size = 100, $m=-2$, $c=0$, $\epsilon \in N(0,0.1)$   & 0.1 & 0.01 & 0.02 & 0.084 & 5.346 \\

\bottomrule
\end{tabular}

\label{tab: ACSVR_100}
\end{table}

\begin{table}[htbp]
\centering
\caption{\textbf{Hyperparameters and Performance metrics of Augmented Chaotic SVR on dataset of the form $\{(x_i,y_i):y_i=mx_i+c+\epsilon,i=1,2,\ldots,n\}$ with sample size $n=1000$.}}

\begin{tabular}{lcccccc}
\toprule
\textbf{Dataset Parameters} & \textbf{alpha} & \textbf{q} & \textbf{epsilon} & \textbf{Training MSE} & \textbf{Testing MSE} \\
\midrule
Size = 1000, $m=2$, $c=0$, $\epsilon \in N(0,5)$      & 0.1 & 0.02 & 0.01 & 4.799 & 4.778 \\
Size = 1000, $m=-2$, $c=0$, $\epsilon \in N(0,5)$     & 0.1 & 0.01 & 0.01 & 4.829 & 4.827 \\
Size = 1000, $m=2$, $c=0$, $\epsilon \in N(0,1)$      & 0.1 & 0.01 & 0.01 & 0.958 & 0.989 \\
Size = 1000, $m=-2$, $c=0$, $\epsilon \in N(0,1)$     & 0.1 & 0.01 & 0.01 & 0.973 & 1.014 \\
Size = 1000, $m=2$, $c=0$, $\epsilon \in N(0,0.1)$    & 0.1 & 0.01 & 0.01 & 0.098 & 0.144 \\
Size = 1000, $m=-2$, $c=0$, $\epsilon \in N(0,0.1)$   & 0.1 & 0.01 & 0.01 & 0.102 & 0.149 \\
\bottomrule
\end{tabular}

\label{tab: ACSVR_1000}
\end{table}

\section{Comparative Analysis}

To evaluate the effectiveness of the proposed augmented chaotic regression methods, a comparative analysis was conducted using four models: Linear Regression (LR) , Ridge Regression (RR), Lasso Regression (LS), and Support Vector Regression (SVR), along with their respective augmented counterparts. The $R^2$ scores across ten real-world datasets were compared to identify performance trends and insights.

The following key observations were made:
\begin{itemize}
    \item Across almost all datasets, the augmented versions consistently outperform or match their base methods, demonstrating the potential of the augumentation using chaotic features. In particular, Support Vector Regression (SVR) and its augmented version exhibit the highest $R^2$ Scores across several datasets, indicating superior modeling capacity for nonlinear patterns.
    \item The Bike Sharing Dataset consistently achieves very high $R^2$ scores across all models. Similarly, the Fish Dataset yields high $R^2$ Score values for all methods, with the ACSVR achieving the highest $(0.97)$.
    \item Algorithm with high $R^2$ score has better performance. Boost in $R^2$ Score is computed using 
     $$\text{Boost in $R^2 Score$} = \frac{R^2_{\text{Augumented Chaotic Regression}}-R^2_{Traditional Regression}}{R^2_{Traditional Regression}}.$$ A dataset is considered "improved" if the boost value is positive. Average boost is calculated only over the improved datasets for each algorithm. The table \ref{tab:real aug boost-summary}  shows how many datasets show improvement for each of the four algorithms when using the augmented version. Along with the count, the average boost percentage across those improved datasets is provided. Tables \ref{tab:LR real}, \ref{tab:RR real}, \ref{tab:LS real}, \ref{tab:SVR real} presents performance metrics of Traditional regression algorithms.

     \item Moderate $R^2$ scores are observed for all models, indicating the dataset’s complexity. Notably, Lasso Regression required careful tuning of the regularization parameter to avoid near-zero $R^2$ scores, highlighting its sensitivity.
\end{itemize}

\begin{table}[htbp]
\centering
\caption{\textbf{Number of real life datasets with Improvement and Average Boost Percentage for Augumented Chaotic Regression Algorithms.}}
\begin{tabular}{lcc}
\toprule
\textbf{Algorithm} & \textbf{No. of Datasets Improved} & \textbf{Average Boost (\%)} \\
\midrule
ACLR & 4 & 8.92\% \\
ACLS & 6 & 10.24\% \\
ACRR & 5 & 11.35\% \\
ACSVR & 6 & 4.95\% \\
\bottomrule
\end{tabular}

\label{tab:real aug boost-summary}
\end{table}

Corresponding bar graphs for each model comparison have been included to visually support the above analysis.

\begin{figure*}[htbp]
    \centering
    \includegraphics[width=0.90\linewidth]{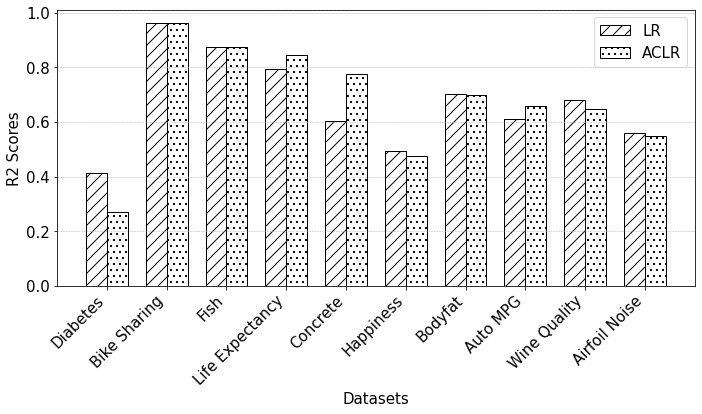}
    \caption{Comparison of $R^2$ Scores between LR and ACLR across Real-World Datasets.}
    \label{LR vs ACLR real}
\end{figure*}

\begin{figure*}[htbp]
    \centering
    \includegraphics[width=0.90\linewidth]{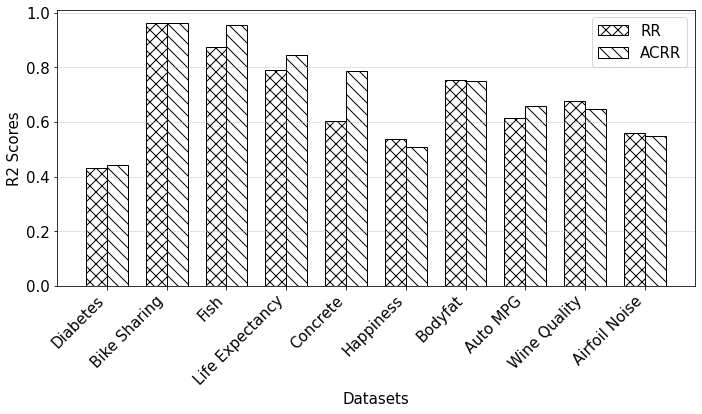}
    \caption{Comparison of $R^2$ Scores between RR and ACRR across Real-World Datasets.}
    \label{RR vs ACRR real}
\end{figure*}

\begin{figure*}[htbp]
    \centering
    \includegraphics[width=0.90\linewidth]{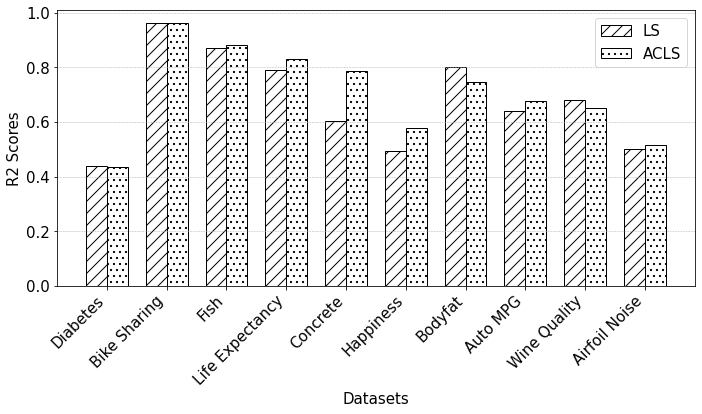}
    \caption{Comparison of $R^2$ Scores between LS and ACLS across Real-World Datasets.}
    \label{LS vs ACLS real}
\end{figure*}

\begin{figure*}[htbp]
    \centering
    \includegraphics[width=0.90\linewidth]{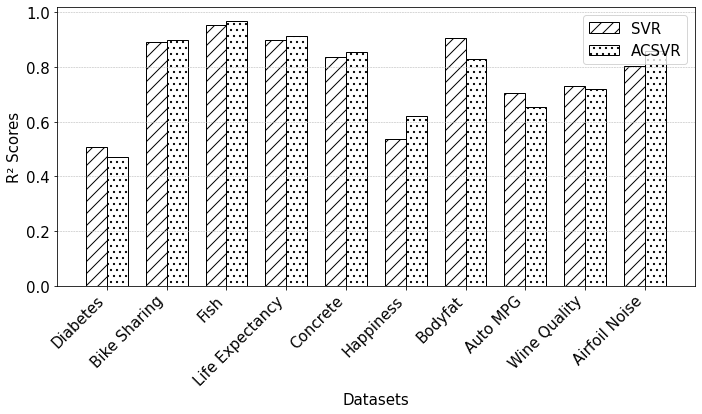}
    \caption{Comparison of $R^2$ Scores between SVR and ACSVR across Real-World Datasets.}
    \label{SVr vs ACSVR real}
\end{figure*}

\begin{figure*}[htbp]
    \centering
    \includegraphics[width=0.90\linewidth]{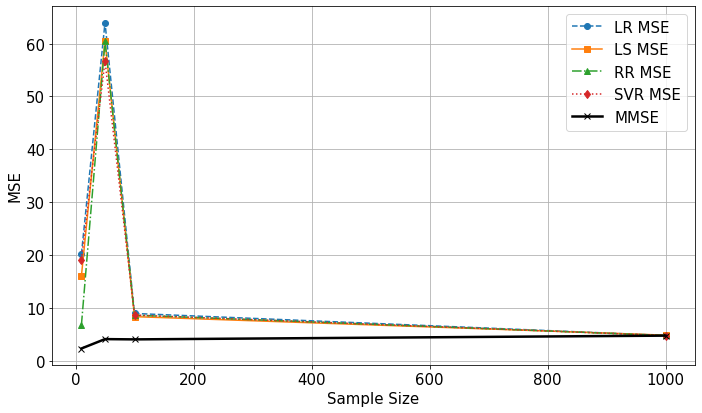}
    \caption{Comparison of MSE of traditional regression methods and MMSE corresponding to the generated datasets of different size.}
    \label{TraditionalMSE and MMSE}
\end{figure*}

\begin{figure*}[htbp]
    \centering
    \includegraphics[width=0.90\linewidth]{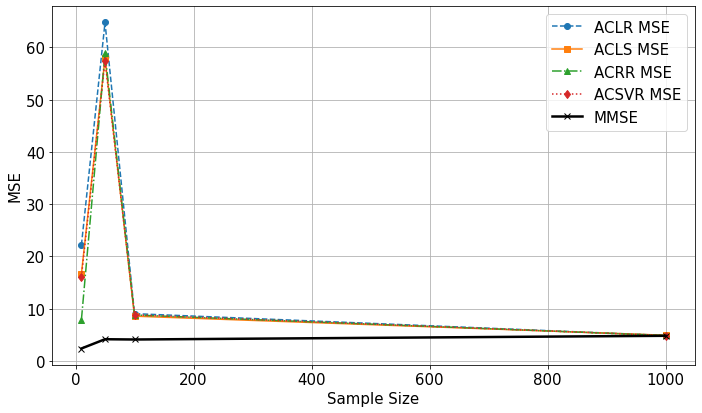}
    \caption{Comparison of MSE of Augmented chaotic regression methods and MMSE corresponding to the generated datasets of different size.}
    \label{AugmentedMSE and MMSE}
\end{figure*}

To compare the performance of the proposed algorithms, MSE and MMSE of generated datasets of different sizes are also compared.
\begin{itemize}
\item The figures \ref{TraditionalMSE and MMSE} and \ref{AugmentedMSE and MMSE} shows that MSE and MMSE corresponding to the dataset $\{(x_i,y_i):y_i=2x_i+\epsilon,i=1,2,\ldots,n, \epsilon \in N(0,5)\}$ for $n=10, 50, 100, 1000$ with LR, LS, RR and SVR and their augmented versions. It is clear that as sample size increases, MSE decreases and getting closer to MMSE. 
\item The Table \ref{tab:MMSE_MSE_comparison1000} higlights the MSE that is more close to MMSE corresponding to the generated datasets of size $1000$ under different noise levels.
\item Lasso regression (LS) may produce MSE values lower than the MMSE for certain datasets (For example, for the generated dataset with $Size=1000$, $m=2$, $c=0$, $ \epsilon  \in N(0,5)$ ) due to the effect of the regularisation term, which can lead to underfitting or overfitting depending on the chosen regularisation parameter.Ridge (RR) typically doesn't show this behavior as much because it doesn't eliminate features, making it less likely to underfit or overfit as sharply as Lasso does with poor regularisation values.
\end{itemize}

 \begin{table}[h]
\centering
\caption{\textbf{Comparison of MMSE and MSE values for different regression models on generated datasets of size $1000$.}}
  \resizebox{\textwidth}{!}{ 
\begin{tabular}{lccccccccc}
\toprule
\textbf{Parameters to generate dataset} & \textbf{MMSE} & \textbf{SVR MSE} & \textbf{ASVR MSE} & \textbf{LR MSE} & \textbf{ACLR MSE} & \textbf{RR MSE} & \textbf{ACRR MSE} & \textbf{LS MSE} & \textbf{ACLS MSE} \\
\midrule
$Size=1000$, $m=2$, $c=0$, $ \epsilon  \in N(0,5)$ & 4.783 & 4.788 & 4.778 & 4.781 & 4.795 & 4.775 & 4.792 & 4.792 &\textbf{ 4.755} \\
$Size=1000$, $m=-2$, $c=0$, $ \epsilon  \in N(0,5)$ & 4.783 & \textbf{4.806} & 4.827 & 4.815 & 4.847 & 4.812 & 4.841 & 4.871 & 4.888 \\
$Size=1000$, $m=2$, $c=0$, $ \epsilon  \in N(0,1)$ & 0.957 & 0.994 & 0.989 & 0.990 & 0.985 & 0.985 & \textbf{0.980} & 1.014 & 1.012 \\
$Size=1000$, $m=-2$, $c=0$, $ \epsilon  \in N(0,1)$ & 0.957 & \textbf{1.001} & 1.014 & 1.006 & 1.018 & 1.002 & 1.013 & 1.049 & 1.059 \\
$Size=1000$, $m=2$, $c=0$, $ \epsilon  \in N(0,1)$ & 0.095 & 0.142 & 0.144 & 0.141 & 0.139 & 0.136 & \textbf{0.135} & 0.171 & 0.173 \\
$Size=1000$, $m=-2$, $c=0$, $ \epsilon  \in N(0,1)$ & 0.095 & 0.144 & 0.149 & 0.145 & 0.150 & \textbf{0.141} & 0.145 & 0.182 & 0.188 \\
\bottomrule
\end{tabular}}

\label{tab:MMSE_MSE_comparison1000}
\end{table}

\section{Conclusion}
In this study, the performance of four augmented regression algorithms—Augmented Linear Regression, Augmented Ridge Regression, Augmented Lasso Regression, and Augmented Support Vector Regression (SVR)—was evaluated using ten different real-life datasets and a synthetically generated dataset of the form $\{(x_i,y_i):y_i=mx_i+c+\epsilon,i=1,2,\ldots,n\}$. Our experimental analysis revealed that the inclusion of the chaotic Tracemean feature as an additional input significantly improved the performance of these regression models. Specifically, six out of the ten real-life datasets exhibited improved performance when the chaotic Tracemean feature was incorporated in both Augmented Lasso Regression and Augmented SVR, demonstrating the potential of chaos-inspired features in enhancing model accuracy.

Among the four augmented algorithms, Augmented Chaotic Ridge Regression exhibited the highest average performance boost, achieving an improvement of 11.35\% compared to its non-augmented counterpart. Additionally, experiments conducted on the generated dataset showed that as the sample size increased, the Mean Squared Error (MSE) of all four augmented models consistently decreased and eventually converged towards the Minimum Mean Squared Error (MMSE), further validating the stability and scalability of the proposed approach.

Despite the promising results, this study is not without limitations. One of the primary challenges encountered was the need for extensive hyperparameter tuning, which can be time-consuming and may limit the practical applicability of the models in real-world scenarios.

Future work can focus on two key directions: first, exploring the inclusion of additional chaotic features to further enhance model performance, and second, developing efficient strategies to reduce training time, potentially through hyperparameter optimization techniques or parameter-free approaches.

\clearpage

\begin{appendices}
\section*{Appendix}
\section{Dataset Description}\label{regression_datasets}
\begin{table}[h]
    \centering
    \caption{\textbf{Dataset Details for Regression Analysis.}}
    \begin{tabular}{clcc}
        \toprule
        \textbf{Sl. No} & \textbf{Dataset} & \textbf{Number of Samples} & \textbf{Number of Features} \\
        \midrule
        1  & Diabetes Dataset     & 442  & 10  \\
        2  & Bike Sharing Dataset & 731  & 13  \\
        3  & Fish                & 159  & 5   \\
        4  & Life Expectancy     & 1649 & 18  \\
        5  & Concrete Strength           & 1030 & 8   \\
        6  & World Happiness Report          & 156  & 6   \\
        7  & Body Fat             & 252  & 14  \\
        8  & Auto MPG            & 392  & 5   \\
        9  & Red Wine Quality        & 1599 & 10  \\
        10 & NASA Airfoil Self-Noise       & 1503 & 5   \\
        \bottomrule
    \end{tabular}
    
    \label{reg:dataset_details}
\end{table}
\subsection{Diabetes Dataset}
The Scikit-learn Diabetes Dataset (also known as the Sklearn Diabetes dataset)~\cite{lichman2013uci} includes ten features , such as age, sex, body mass index (BMI), average blood pressure, and six blood serum measurements, collected from 442 diabetes patients. The target variable represents a quantitative assessment of disease progression one year after the baseline measurements.
\subsection{Bike Sharing Dataset}
Bike Sharing dataset~\cite{bike_sharing_275} contains the hourly and daily count of rental bikes from 2011 to 2012 in Capital bikeshare system with the corresponding weather and seasonal information.
\subsection{Fish Dataset}
The fish market dataset ~\cite{mark_daniel_lampa_rose_claire_librojo_mary_mae_calamba_2022} is a collection of data related to various species of fish and their characteristics. The dataset is structured in such a way that each row corresponds to a single fish with its species and various physical measurements (lengths, height, and width).  The target variable represents the weight of a fish based on its species and the provided physical measurements.
\subsection{Life Expectancy Dataset}
 The Life Expectancy dataset~\cite{lasha_gochiashvili_2023} consists of 22 Columns and 2938 rows , including 20 predictive variables. These variables are divided into several broad categories: Immunization related factors, Mortality factors, Economical factors and Social factors. The target variable represents the life expectancy of people in various countries of the world. 
 
\subsection{Concrete Strength Dataset}
The Concrete Dataset~\cite{concrete_compressive_strength_165} provides essential information on concrete mixtures, including ingredient composition and strength characteristics.The target variable represents the compressive strength of the concrete.

\subsection{World Happiness Report}
The Happiness Dataset~\cite{world_happiness_kaggle} includes a happiness score and six key contributing factors: economic production, social support, life expectancy, freedom, absence of corruption, and generosity. The happiness score estimates the extent to which each of these six factors contributes to making life evaluations higher in each country than they are in Dystopia, a hypothetical country that has values equal to the world’s lowest national averages for each of the six factors. The target variable represents happiness score.

\subsection{Body Fat Dataset}
The Body Fat Dataset~\cite{body_fat_kaggle} includes measurement of body fat, age, weight, height, neck circumference, Knee circumference, ankle circumference, etc. The target variable represents the measurement of body fat.\

\subsection{Auto MPG Dataset}
The Auto MPG Dataset~\cite{auto_mpg_9} is on city-cycle fuel consumption measured in miles per gallon (mpg). The dataset includes Car name, model year, weight, cylinders, etc. The target variable represents Mileage per gallon performances of various cars.

\subsection{Red Wine Quality Dataset}
The Wine Quality Dataset~\cite{wine_quality_186} contains physicochemical properties of Portuguese "Vinho Verde" red wines and their corresponding quality scores, ranging from 0 to 10. It includes 11 input features such as acidity, alcohol content, pH, sulphates, and sugar levels, which influence the sensory evaluation of wine. The target variable represents the quality of wine. 

\subsection{NASA Airfoil Self-Noise Dataset}
The NASA Airfoil Self-Noise Dataset~\cite{airfoil_self-noise_291} contains aerodynamic and acoustic test data collected from NACA 0012 airfoil blade sections in an anechoic wind tunnel. It includes five input features—frequency, angle of attack, chord length, free-stream velocity, and suction side displacement thickness—used to predict the scaled sound pressure level (SSPL) in decibels.

\section{Performance Analysis of traditional regression methods}
This section presents the value of performance metrics of traditional regression methods including Linear Regression, Ridge Regression, Lasso Regression and Support Vector Regression. The performance is analysed on ten distict real life datasets described in section \ref{regression_datasets} and on the generated dataset of the form $\{(x_i,y_i):y_i=mx_i+c+\epsilon,i=1,2,\ldots,n\}$.
\subsection{Linear Regression(LR)}
Linear Regression does not involve any tunable hyperparameters. The table \ref{tab:LR real} presents the value of hyperparameters and the performance metrics $R^2$ Score, MSE and MAE for real world datasets. Tables \ref{tab:LR 10}, \ref{tab:LR 50}, \ref{tab:LR 100}, \ref{tab:LR 1000} correseponds to the dataset of size $10$, $50$, $100$ and $1000$ respectively
\begin{table}[htbp]
    \centering
    \caption{\textbf{MSE of LR on real life datasets.}}
    \begin{tabular}{cllll}
\toprule
\textbf{Sl. No} & \textbf{Dataset} & \textbf{R\textsuperscript{2} Score} & \textbf{MSE} & \textbf{MAE} \\
\midrule
1  & Diabetes            & 0.413 & 3109.163   & 43.485 \\
2  & Bike Sharing        & 0.961 & 154179.697 & 373.843 \\
3  & Fish                & 0.875 & 17662.023  & 102.336 \\
4  & Life Expectancy     & 0.792 & 14.762     & 2.904 \\
5  & Concrete            & 0.604 & 113.229    & 8.426 \\
6  & Happiness           & 0.492 & 0.528      & 0.529 \\
7  & Bodyfat             & 0.704 & 13.773     & 2.970 \\
8  & Auto MPG            & 0.611 & 19.859     & 3.740 \\
9  & Wine Quality        & 0.679 & 0.400      & 0.494 \\
10 & Airfoil Noise       & 0.558 & 22.128     & 3.672 \\
\bottomrule
\end{tabular}
    
    \label{tab:LR real}
\end{table}

\begin{table}[htbp]
    \centering
     \caption{\textbf{MSE of LR on generated dataset of size 10.}}
   \begin{tabular}{ll}
\toprule
\textbf{Parameters to generate dataset } & \textbf{LR MSE} \\
\midrule
$\text{Size} = 10$, $m = 2$, $c=0$, $\epsilon\in N(0,5)$     & 20.014 \\
$\text{Size} = 10$, $m = -2$, $c=0$, $\epsilon\in N(0,5)$    & 20.192 \\
$\text{Size} = 10$, $m = 2$, $c=0$, $\epsilon\in N(0,1)$     & 22.560 \\
$\text{Size} = 10$, $m = -2$, $c=0$, $\epsilon\in N(0,1)$    & 18.802 \\
$\text{Size} = 10$, $m = 2$, $c=0$, $\epsilon\in N(0,0.1)$   & 20.440 \\
$\text{Size} = 10$, $m = -2$, $c=0$, $\epsilon\in N(0,0.1)$  & 19.251 \\
\bottomrule
\end{tabular}
   
    \label{tab:LR 10}
\end{table}

\begin{table}[htbp]
    \centering
     \caption{\textbf{MSE of LR on generated dataset of size 50.}}
    \begin{tabular}{ll}
\toprule
\textbf{Parameters to generate dataset} & \textbf{LR MSE} \\
\midrule
$\text{Size} = 50$, $m = 2$, $c=0$, $\epsilon\in N(0,5)$     & 27.524 \\
$\text{Size} = 50$, $m = -2$, $c=0$, $\epsilon\in N(0,5)$    & 63.974 \\
$\text{Size} = 50$, $m = 2$, $c=0$, $\epsilon\in N(0,1)$     & 33.775 \\
$\text{Size} = 50$, $m = -2$, $c=0$, $\epsilon\in N(0,1)$    & 50.076 \\
$\text{Size} = 50$, $m = 2$, $c=0$, $\epsilon\in N(0,0.1)$   & 38.487 \\
$\text{Size} = 50$, $m = -2$, $c=0$, $\epsilon\in N(0,0.1)$  & 43.643 \\
\bottomrule
\end{tabular}
   
    \label{tab:LR 50}
\end{table}

\begin{table}[htbp]
    \centering
     \caption{\textbf{MSE of LR on generated dataset of size 100.}}
    \begin{tabular}{ll}
\toprule
\textbf{Parameters to generate dataset} & \textbf{LR MSE} \\
\midrule
$\text{Size} = 100$, $m = 2$, $c=0$, $\epsilon\in N(0,5)$     & 7.321 \\
$\text{Size} = 100$, $m = -2$, $c=0$, $\epsilon\in N(0,5)$    & 8.972 \\
$\text{Size} = 100$, $m = 2$, $c=0$, $\epsilon\in N(0,1)$     & 5.294 \\
$\text{Size} = 100$, $m = -2$, $c=0$, $\epsilon\in N(0,1)$    & 6.033 \\
$\text{Size} = 100$, $m = 2$, $c=0$, $\epsilon\in N(0,0.1)$   & 4.988 \\
$\text{Size} = 100$, $m = -2$, $c=0$, $\epsilon\in N(0,0.1)$  & 5.221 \\
\bottomrule
\end{tabular}
   
    \label{tab:LR 100}
\end{table}

\begin{table}[htbp]
    \centering
    \caption{\textbf{MSE of LR on generated dataset of size 1000.}}
   \begin{tabular}{ll}
\toprule
\textbf{Parameters to generate dataset} & \textbf{LR MSE} \\
\midrule
$\text{Size} = 1000$, $m = 2$, $c=0$, $\epsilon\in N(0,5)$     & 4.781 \\
$\text{Size} = 1000$, $m = -2$, $c=0$, $\epsilon\in N(0,5)$    & 4.815 \\
$\text{Size} = 1000$, $m = 2$, $c=0$, $\epsilon\in N(0,1)$     & 0.990 \\
$\text{Size} = 1000$, $m = -2$, $c=0$, $\epsilon\in N(0,1)$    & 1.006 \\
$\text{Size} = 1000$, $m = 2$, $c=0$, $\epsilon\in N(0,0.1)$   & 0.141 \\
$\text{Size} = 1000$, $m = -2$, $c=0$, $\epsilon\in N(0,0.1)$  & 0.145 \\
\bottomrule
\end{tabular}
    
    \label{tab:LR 1000}
\end{table}

\subsection{Ridge Regression(RR)}
The regularisation parameter $\alpha$ in  Ridge Regression is tuned among the values $0.1,1.0 \text{ and } 10.0$. The table \ref{tab:RR real} presents the value of hyperparameters and the performance metrics $R^2$ Score, MSE and MAE for real world datasets. Tables \ref{tab:RR 10}, \ref{tab:RR 50}, \ref{tab:RR 100}, \ref{tab:RR 1000} correseponds to the dataset of size $10$, $50$, $100$ and $1000$ respectively

\begin{table}[htbp]
    \centering
     \caption{\textbf{MSE of RR on real life datasets.}}
   
    \begin{tabular}{clcccll}
\toprule
\textbf{Sl. No.} & \textbf{Dataset} & \boldmath$\alpha$ & \textbf{Training $R^2$ Score} & \textbf{Testing $R^2$ Score} & \textbf{MSE} & \textbf{MAE} \\
\midrule
1 & Diabetes Dataset     & 0.1 & 0.480 & 0.430 & 3022.264   & 43.083  \\
2 & Bike Sharing Dataset & 0.1 & 0.999 & 0.962 & 153219.576 & 372.521 \\
3 & Fish                 & 0.1 & 0.874 & 0.874 & 17943.463  & 105.318 \\
4 & Life Expectancy      & 0.1 & 0.826 & 0.789 & 14.969     & 2.929   \\
5 & Concrete             & 0.1 & 0.602 & 0.605 & 112.990    & 8.431   \\
6 & Happiness            & 1.0 & 0.784 & 0.538 & 0.480      & 0.517   \\
7 & Bodyfat              & 0.1 & 0.967 & 0.752 & 11.518     & 2.691   \\
8 & AutoMPG              & 0.1 & 0.706 & 0.613 & 19.744     & 3.722   \\
9 & Wine Quality         & 0.1 & 0.647 & 0.678 & 0.401      & 0.497   \\
10 & Airfoil Noise       & 0.1 & 0.492 & 0.558 & 22.142     & 3.675   \\
\bottomrule
\end{tabular}
   
    \label{tab:RR real}
\end{table}

\begin{table}[htbp]
    \centering
      \caption{\textbf{MSE of RR on generated dataset of size 10.}}
    \begin{tabular}{lcc}
\toprule
\textbf{Parameters to generate dataset} & \textbf{Training MSE} & \textbf{Testing MSE} \\
\midrule
$\text{Size} = 10$, $m$ = 2, $c$ = 0, $\epsilon\in N(0,5)$      & 12.240 & 10.438 \\
$\text{Size} = 10$, $m$ = -2, $c$ = 0, $\epsilon\in N(0,5)$     & 11.533 & 6.768  \\
$\text{Size} = 10$, $m$ = 2, $c$ = 0, $\epsilon\in N(0,1)$      & 9.885  & 5.720  \\
$\text{Size} = 10$, $m$ = -2, $c$ = 0, $\epsilon\in N(0,1)$     & 9.568  & 4.079  \\
$\text{Size} = 10$, $m$ = 2, $c$ = 0, $\epsilon\in N(0,0.1)$    & 9.291  & 4.326  \\
$\text{Size} = 10$, $m$ = -2, $c$ = 0, $\epsilon\in N(0,0.1)$   & 9.191  & 3.807  \\
\bottomrule
    \end{tabular}
  
    \label{tab:RR 10}
\end{table}

\begin{table}[htbp]
    \centering
    \caption{\textbf{MSE of RR on generated dataset of size 50.}}
    \begin{tabular}{lcc}
\toprule
\textbf{Parameters to generate dataset $y = mx + c + \text{Noise}$} & \textbf{Training MSE} & \textbf{Testing MSE} \\
\midrule
$\text{Size} = 50$, $m$ = 2, $c$ = 0, $\epsilon\in N(0,5)$      & 4.899 & 26.144 \\
$\text{Size} = 50$, $m$ = -2, $c$ = 0, $\epsilon\in N(0,5)$     & 4.800 & 60.425 \\
$\text{Size} = 50$, $m$ = 2, $c$ = 0, $\epsilon\in N(0,1)$      & 1.131 & 31.831 \\
$\text{Size} = 50$, $m$ = -2, $c$ = 0, $\epsilon\in N(0,1)$     & 1.086 & 47.162 \\
$\text{Size} = 50$, $m$ = 2, $c$ = 0, $\epsilon\in N(0,0.1)$    & 0.273 & 36.220 \\
$\text{Size} = 50$, $m$ = -2, $c$ = 0, $\epsilon\in N(0,0.1)$   & 0.259 & 41.068 \\
\bottomrule
\end{tabular}
    
    \label{tab:RR 50}
\end{table}

\begin{table}[htbp]
    \centering
    \caption{\textbf{MSE of RR on generated dataset of size 100.}}
    \begin{tabular}{lcc}
\toprule
\textbf{Parameters to generate dataset} & \textbf{Training MSE} & \textbf{Testing MSE} \\
\midrule
$\text{Size} = 100$, $m$ = 2, $c$ = 0, $\epsilon\in N(0,5)$      & 4.497 & 7.047 \\
$\text{Size} = 100$, $m$ = -2, $c$ = 0, $\epsilon\in N(0,5)$     & 4.513 & 8.649 \\
$\text{Size} = 100$, $m$ = 2, $c$ = 0, $\epsilon\in N(0,1)$      & 0.932 & 5.006 \\
$\text{Size} = 100$, $m$ = -2, $c$ = 0, $\epsilon\in N(0,1)$     & 0.940 & 5.723 \\
$\text{Size} = 100$, $m$ = 2, $c$ = 0, $\epsilon\in N(0,0.1)$    & 0.132 & 4.692 \\
$\text{Size} = 100$, $m$ = -2, $c$ = 0, $\epsilon\in N(0,0.1)$   & 0.134 & 4.919 \\
\bottomrule
\end{tabular}
    
    \label{tab:RR 100}
\end{table}

\begin{table}[htbp]
    \centering
    \caption{\textbf{MSE of RR on generated dataset of size 1000.}}
   \begin{tabular}{lcc}
\toprule
\textbf{Parameters to generate dataset } & \textbf{Training MSE} & \textbf{Testing MSE} \\
\midrule
$\text{Size} = 1000$, $m$ = 2, $c$ = 0, $\epsilon\in N(0,5)$      & 4.805 & 4.775 \\
$\text{Size} = 1000$, $m$ = -2, $c$ = 0, $\epsilon\in N(0,5)$     & 4.805 & 4.812 \\
$\text{Size} = 1000$, $m$ = 2, $c$ = 0, $\epsilon\in N(0,1)$      & 0.961 & 0.985 \\
$\text{Size} = 1000$, $m$ = -2, $c$ = 0, $\epsilon\in N(0,1)$     & 0.961 & 1.002 \\
$\text{Size} = 1000$, $m$ = 2, $c$ = 0, $\epsilon\in N(0,0.1)$    & 0.096 & 0.136 \\
$\text{Size} = 1000$, $m$ = -2, $c$ = 0, $\epsilon\in N(0,0.1)$   & 0.096 & 0.141 \\
\bottomrule
\end{tabular}
    
    \label{tab:RR 1000}
\end{table}

\subsection{Lasso Regression(LS)}
The regularisation parameter $\alpha$ in Lasso Regression and Ridge Regression is tuned among the values $0.1,1.0 \text{ and } 10.0$. The maximum number of iterations used in Lasso Regression is $10000$. The table \ref{tab:LS real} presents the value of hyperparameters and the performance metrics $R^2$ Score, MSE and MAE for real world datasets. Tables \ref{tab:LS 10}, \ref{tab:LS 50}, \ref{tab:LS 100}, \ref{tab:LS 1000} correseponds to the dataset of size $10$, $50$, $100$ and $1000$ respectively

\begin{table}[htbp]
    \centering
    \caption{\textbf{MSE of LS on real life datasets.}}
    
    \begin{tabular}{cccccll}
\toprule
\textbf{Sl. No.} & \textbf{Dataset} & \boldmath$\alpha$ & \textbf{Training $R^2$ Score} & \textbf{Testing $R^2$ Score} & \textbf{MSE} & \textbf{MAE} \\
\midrule
1 & Diabetes Dataset     & 0.1    & 0.481 & 0.439 & 2968.989 & 42.904 \\
2 & Bike Sharing Dataset & 0.1    & 0.999 & 0.961 & 154087.428 & 373.799 \\
3 & Fish                 & 0.1    & 0.875 & 0.872 & 18104.369 & 104.990 \\
4 & Life Expectancy      & 0.1    & 0.805 & 0.790 & 14.899 & 2.875 \\
5 & Concrete             & 0.1    & 0.596 & 0.603 & 113.502 & 8.471 \\
6 & Happiness            & 0.1    & 0.534 & 0.495 & 0.525 & 0.565 \\
7 & Bodyfat              & 0.1    & 0.965 & 0.802 & 9.207 & 2.444 \\
8 & Auto MPG             & 0.1    & 0.704 & 0.641 & 18.306 & 3.494 \\
9 & Wine Quality \textsuperscript{*}        & 0.0001 & 0.647 & 0.680 & 0.397 & 0.493 \\
10 & Airfoil Noise       & 0.1    & 0.461 & 0.502 & 24.910 & 3.999 \\
\bottomrule
\end{tabular}
\vspace{1mm}
\begin{flushleft}
\textsuperscript{*}\textit{Note:} Unlike the other datasets, the Wine quality dataset required tuning of $\alpha$ from the smaller values 0.0001, 0.001, and 0.01. Larger values such as 0.1, 1, and 10 resulted in an $R^2$ Score of 0 for standalone Lasso Regression.
\end{flushleft}
    
    \label{tab:LS real}
    
\end{table}

\begin{table}[htbp]
    \centering
     \caption{\textbf{MSE of LS on generated dataset of size 10.}}
    \begin{tabular}{llll}
\toprule
\textbf{Parameters to generate dataset} & \boldmath$\alpha$ & \textbf{Training MSE} & \textbf{Testing MSE} \\
\midrule
$\text{Size} = 10$, $m = 2$, $c=0$, $\epsilon\in N(0,5)$     & 0.1 & 3.139 & 23.413 \\
$\text{Size} = 10$, $m = -2$, $c=0$, $\epsilon\in N(0,5)$    & 0.1 & 3.096 & 16.066 \\
$\text{Size} = 10$, $m = 2$, $c=0$, $\epsilon\in N(0,1)$     & 0.1 & 0.921 & 17.670 \\
$\text{Size} = 10$, $m = -2$, $c=0$, $\epsilon\in N(0,1)$    & 0.1 & 0.901 & 14.384 \\
$\text{Size} = 10$, $m = 2$, $c=0$, $\epsilon\in N(0,0.1)$   & 0.1 & 0.418 & 15.711 \\
$\text{Size} = 10$, $m = -2$, $c=0$, $\epsilon\in N(0,0.1)$  & 0.1 & 0.412 & 14.672 \\
\bottomrule
\end{tabular}
   
    \label{tab:LS 10}
\end{table}

\begin{table}[htbp]
    \centering
    \caption{\textbf{MSE of LS on generated dataset of size 50.}}
    \begin{tabular}{llll}
\toprule
\textbf{Parameters to generate dataset} & \boldmath$\alpha$ & \textbf{Training MSE} & \textbf{Testing MSE} \\
\midrule
$\text{Size} = 50$, $m = 2$, $c=0$, $\epsilon\in N(0,5)$     & 0.1 & 4.858 & 26.046 \\
$\text{Size} = 50$, $m = -2$, $c=0$, $\epsilon\in N(0,5)$    & 0.1 & 4.750 & 60.487 \\
$\text{Size} = 50$, $m = 2$, $c=0$, $\epsilon\in N(0,1)$     & 0.1 & 1.079 & 31.742 \\
$\text{Size} = 50$, $m = -2$, $c=0$, $\epsilon\in N(0,1)$    & 0.1 & 1.031 & 47.145 \\
$\text{Size} = 50$, $m = 2$, $c=0$, $\epsilon\in N(0,0.1)$   & 0.1 & 0.220 & 36.148 \\
$\text{Size} = 50$, $m = -2$, $c=0$, $\epsilon\in N(0,0.1)$  & 0.1 & 0.204 & 41.018 \\
\bottomrule
\end{tabular}
    
    \label{tab:LS 50}
\end{table}

\begin{table}[htbp]
    \centering
    \caption{\textbf{MSE of LS on generated dataset of size 100.}}
    \begin{tabular}{llll}
\toprule
\textbf{Parameters to generate dataset} & \boldmath$\alpha$ & \textbf{Training MSE} & \textbf{Testing MSE} \\
\midrule
$\text{Size} = 100$, $m = 2$, $c=0$, $\epsilon\in N(0,5)$     & 0.1 & 4.568 & 6.831 \\
$\text{Size} = 100$, $m = -2$, $c=0$, $\epsilon\in N(0,5)$    & 0.1 & 4.597 & 8.374 \\
$\text{Size} = 100$, $m = 2$, $c=0$, $\epsilon\in N(0,1)$     & 0.1 & 1.006 & 4.774 \\
$\text{Size} = 100$, $m = -2$, $c=0$, $\epsilon\in N(0,1)$    & 0.1 & 1.018 & 5.464 \\
$\text{Size} = 100$, $m = 2$, $c=0$, $\epsilon\in N(0,0.1)$   & 0.1 & 0.207 & 4.451 \\
$\text{Size} = 100$, $m = -2$, $c=0$, $\epsilon\in N(0,0.1)$  & 0.1 & 0.211 & 4.669 \\
\bottomrule
\end{tabular}
    
    \label{tab:LS 100}
\end{table}

\begin{table}[htbp]
    \centering
     \caption{\textbf{MSE of LS on generated dataset of size 1000.}}
    \begin{tabular}{llll}
\toprule
\textbf{Parameters to generate dataset} & \boldmath$\alpha$ & \textbf{Training MSE} & \textbf{Testing MSE} \\
\midrule
$\text{Size} = 1000$, $m = 2$, $c=0$, $\epsilon\in N(0,5)$     & 0.1 & 4.923 & 4.792 \\
$\text{Size} = 1000$, $m = -2$, $c=0$, $\epsilon\in N(0,5)$    & 0.1 & 4.926 & 4.871 \\
$\text{Size} = 1000$, $m = 2$, $c=0$, $\epsilon\in N(0,1)$     & 0.1 & 1.080 & 1.014 \\
$\text{Size} = 1000$, $m = -2$, $c=0$, $\epsilon\in N(0,1)$    & 0.1 & 1.081 & 1.049 \\
$\text{Size} = 1000$, $m = 2$, $c=0$, $\epsilon\in N(0,0.1)$   & 0.1 & 0.216 & 0.171 \\
$\text{Size} = 1000$, $m = -2$, $c=0$, $\epsilon\in N(0,0.1)$  & 0.1 & 0.216 & 0.182 \\
\bottomrule
\end{tabular}
   
    \label{tab:LS 1000}
\end{table}

\subsection{Support Vector Regression (SVR)}
The regularisation parameter $C$ in SVR is tuned among the values $1, 10, 50, 100$. The kernel used in real life datasets is radial basis function and is linear in generated dataset. The table \ref{tab:SVR real} presents the value of hyperparameters and the performance metrics $R^2$ Score, MSE and MAE for real world datasets. Tables \ref{tab:SVR 10}, \ref{tab:SVR 50}, \ref{tab:SVR 100}, \ref{tab:SVR 1000} correseponds to the dataset of size $10$, $50$, $100$ and $1000$ respectively

\begin{table}[htbp]
    \centering
    \caption{\textbf{MSE of SVR on real life datasets.}}
    \begin{tabular}{cllcll}
\toprule
\textbf{Sl no} & \textbf{Datasets} & \textbf{C - standalone} & \textbf{Testing R\textsuperscript{2} Score} & \textbf{MSE} & \textbf{MAE} \\
\midrule
1  & Diabetes Dataset      & 50  & 0.508 & 2602.347   & 40.55  \\
2  & Bike Sharing Dataset  & 100 & 0.890 & 440614.581 & 475.00 \\
3  & Fish                  & 100 & 0.952 & 6824.104   & 46.58  \\
4  & Life Expectancy       & 100 & 0.900 & 7.097      & 1.702  \\
5  & Concrete              & 100 & 0.836 & 46.818     & 4.859  \\
6  & Happiness             & 10  & 0.535 & 0.483      & 0.574  \\
7  & Bodyfat               & 50  & 0.907 & 4.325      & 1.702  \\
8  & AutoMPG               & 10  & 0.704 & 15.107     & 2.832  \\
9  & Wine Quality          & 10  & 0.731 & 0.334      & 0.436  \\
10 & Airfoil Noise         & 100 & 0.803 & 9.840      & 2.253  \\
\bottomrule
\end{tabular}
    
    \label{tab:SVR real}
\end{table}

\begin{table}[htbp]
    \centering
     \caption{\textbf{MSE of SVR on generated dataset of size 10.}}
    \begin{tabular}{llll}
\toprule
\textbf{Parameters to generate dataset} & \textbf{C value} & \textbf{Training MSE} & \textbf{Testing MSE} \\
\midrule
$\text{Size} = 10$, $m = 2$, $c=0$, $\epsilon\in N(0,5)$     & 100 & 6.629 & 27.026 \\
$\text{Size} = 10$, $m = -2$, $c=0$, $\epsilon\in N(0,5)$    & 100 & 3.978 & 19.169 \\
$\text{Size} = 10$, $m = 2$, $c=0$, $\epsilon\in N(0,1)$     & 100 & 1.208 & 22.317 \\
$\text{Size} = 10$, $m = -2$, $c=0$, $\epsilon\in N(0,1)$    & 100 & 0.769 & 18.228 \\
$\text{Size} = 10$, $m = 2$, $c=0$, $\epsilon\in N(0,0.1)$   & 100 & 0.096 & 19.643 \\
$\text{Size} = 10$, $m = -2$, $c=0$, $\epsilon\in N(0,0.1)$  & 100 & 0.079 & 18.427 \\
\bottomrule
\end{tabular}
   
    \label{tab:SVR 10}
\end{table}

\begin{table}[htbp]
    \centering
     \caption{\textbf{MSE of SVR on generated dataset of size 50.}}
    \begin{tabular}{llll}
\toprule
\textbf{Parameters to generate dataset} & \textbf{C value} & \textbf{Training MSE} & \textbf{Testing MSE} \\
\midrule
$\text{Size} = 50$, $m = 2$, $c=0$, $\epsilon\in N(0,5)$     & 50  & 5.018  & 31.180 \\
$\text{Size} = 50$, $m = -2$, $c=0$, $\epsilon\in N(0,5)$    & 50  & 4.868  & 56.703 \\
$\text{Size} = 50$, $m = 2$, $c=0$, $\epsilon\in N(0,1)$     & 100 & 0.992  & 35.655 \\
$\text{Size} = 50$, $m = -2$, $c=0$, $\epsilon\in N(0,1)$    & 100 & 0.965  & 47.614 \\
$\text{Size} = 50$, $m = 2$, $c=0$, $\epsilon\in N(0,0.1)$   & 100 & 0.097  & 38.740 \\
$\text{Size} = 50$, $m = -2$, $c=0$, $\epsilon\in N(0,0.1)$  & 100 & 0.0951 & 43.399 \\
\bottomrule
\end{tabular}
   
    \label{tab:SVR 50}
\end{table}

\begin{table}[htbp]
    \centering
     \caption{\textbf{MSE of SVR on generated dataset of size 100.}}
    \begin{tabular}{llll}
\toprule
\textbf{Parameters to generate dataset} & \textbf{C value} & \textbf{Training MSE} & \textbf{Testing MSE} \\
\midrule
$\text{Size} = 100$, $m = 2$, $c=0$, $\epsilon\in N(0,5)$     & 50  & 4.489 & 7.389 \\
$\text{Size} = 100$, $m = -2$, $c=0$, $\epsilon\in N(0,5)$    & 100 & 4.474 & 8.666 \\
$\text{Size} = 100$, $m = 2$, $c=0$, $\epsilon\in N(0,1)$     & 50  & 0.896 & 5.264 \\
$\text{Size} = 100$, $m = -2$, $c=0$, $\epsilon\in N(0,1)$    & 100 & 0.895 & 5.855 \\
$\text{Size} = 100$, $m = 2$, $c=0$, $\epsilon\in N(0,0.1)$   & 50  & 0.088 & 5.033 \\
$\text{Size} = 100$, $m = -2$, $c=0$, $\epsilon\in N(0,0.1)$  & 50  & 0.088 & 5.142 \\
\bottomrule
\end{tabular}
   
    \label{tab:SVR 100}
\end{table}

\begin{table}[htbp]
    \centering
     \caption{\textbf{MSE of SVR on generated dataset of size 1000.}}
    \begin{tabular}{llll}
\toprule
\textbf{Parameters to generate dataset} & \textbf{C value} & \textbf{Training MSE} & \textbf{Testing MSE} \\
\midrule
$\text{Size} = 1000$, $m = 2$, $c=0$, $\epsilon\in N(0,5)$     & 100 & 4.811 & 4.788 \\
$\text{Size} = 1000$, $m = -2$, $c=0$, $\epsilon\in N(0,5)$    & 50  & 4.812 & 4.806 \\
$\text{Size} = 1000$, $m = 2$, $c=0$, $\epsilon\in N(0,1)$     & 100 & 0.960 & 0.994 \\
$\text{Size} = 1000$, $m = -2$, $c=0$, $\epsilon\in N(0,1)$    & 100 & 0.960 & 1.001 \\
$\text{Size} = 1000$, $m = 2$, $c=0$, $\epsilon\in N(0,0.1)$   & 100 & 0.096 & 0.142 \\
$\text{Size} = 1000$, $m = -2$, $c=0$, $\epsilon\in N(0,0.1)$  & 50  & 0.096 & 0.144 \\
\bottomrule
\end{tabular}
   
    \label{tab:SVR 1000}
\end{table}

\section{Minimum Mean Square Error(MMSE) of generated dataset}
The tables \ref{tab:MMSE 10}, \ref{tab:MMSE 50}, \ref{tab:MMSE 100}, \ref{tab: MMSE 1000} corresponds to the MMSE of the dataset of the form $\{(x_i,y_i):y_i=mx_i+c+\epsilon,i=1,2,\ldots,n\}$ for $n=10$, $n=50$, $n=100$, $n=1000$ respectively. Here $\epsilon$ denotes noise and is randomly chosen from a normal distribution with mean zero.

\begin{table}[h]
    \centering
      \caption{\textbf{MMSE on generated dataset of size 10.}}
    \begin{tabular}{ccclcc}
        \toprule
        \textbf{n} & \textbf{m} & \textbf{c} & \textbf{Noise Distribution} & \textbf{MMSE} \\
\midrule
10 & 2   & 0 & $N(0,5)$   & 2.35 \\
10 & -2  & 0 & $N(0,5)$   & 2.35 \\
10 & 2   & 0 & $N(0,1)$   & 0.47 \\
10 & -2  & 0 & $N(0,1)$   & 0.47 \\
10 & 2   & 0 & $N(0,0.1)$ & 0.047 \\
10 & -2  & 0 & $N(0,0.1)$ & 0.047 \\
        \bottomrule
    \end{tabular}
  
    \label{tab:MMSE 10}
\end{table}

\begin{table}[h]
    \centering
     \caption{\textbf{MMSE on generated dataset of size 50.}}
    \begin{tabular}{ccclcc}
\toprule
\textbf{n} & \textbf{m} & \textbf{c} & \textbf{Noise Distribution} & \textbf{MMSE} \\
\midrule
50 & 2   & 0 & $N(0,5)$   & 4.125 \\
50 & -2  & 0 & $N(0,5)$   & 4.125 \\
50 & 2   & 0 & $N(0,1)$   & 0.825 \\
50 & -2  & 0 & $N(0,1)$   & 0.825 \\
50 & 2   & 0 & $N(0,0.1)$ & 0.0825 \\
50 & -2  & 0 & $N(0,0.1)$ & 0.0825 \\
\bottomrule
\end{tabular}
   
    \label{tab:MMSE 50}
\end{table}

\begin{table}[h]
    \centering
    \caption{\textbf{MMSE on generated dataset of size 100.}}
    \begin{tabular}{ccclcc}
\toprule
\textbf{n} & \textbf{m} & \textbf{c} & \textbf{Noise Distribution} & \textbf{MMSE} \\
\midrule
100 & 2   & 0 & $N(0,5)$   & 4.074 \\
100 & -2  & 0 & $N(0,5)$   & 4.074 \\
100 & 2   & 0 & $N(0,1)$   & 0.815 \\
100 & -2  & 0 & $N(0,1)$   & 0.815 \\
100 & 2   & 0 & $N(0,0.1)$ & 0.081 \\
100 & -2  & 0 & $N(0,0.1)$ & 0.081 \\
\bottomrule
\end{tabular}
    
    \label{tab:MMSE 100}
\end{table}

\begin{table}[h!]
    \centering
     \caption{\textbf{MMSE on generated dataset of size 1000.}}
    \begin{tabular}{ccclcc}
\toprule
\textbf{n} & \textbf{m} & \textbf{c} & \textbf{Noise Distribution} & \textbf{MMSE} \\
\midrule
1000 & 2   & 0 & $N(0,5)$   & 4.783 \\
1000 & -2  & 0 & $N(0,5)$   & 4.783 \\
1000 & 2   & 0 & $N(0,1)$   & 0.957 \\
1000 & -2  & 0 & $N(0,1)$   & 0.957 \\
1000 & 2   & 0 & $N(0,0.1)$ & 0.095 \\
1000 & -2  & 0 & $N(0,0.1)$ & 0.095 \\
\bottomrule
\end{tabular}
   
    \label{tab: MMSE 1000}
\end{table}

\end{appendices}

\clearpage
\bibliographystyle{unsrt}  
\bibliography{main}

\end{document}